\documentclass{article} 
\usepackage{iclr2026_conference,times}

\usepackage{amsmath,amsfonts,bm}

\def\eqref#1{equation~\ref{#1}}

\def\1{\bm{1}}

\DeclareMathAlphabet{\mathsfit}{\encodingdefault}{\sfdefault}{m}{sl}
\SetMathAlphabet{\mathsfit}{bold}{\encodingdefault}{\sfdefault}{bx}{n}

\usepackage{graphicx}
\usepackage{amsmath}
\usepackage{float}

\usepackage{hyperref}
\usepackage{url}
\usepackage{cleveref}
\PassOptionsToPackage{numbers, compress}{natbib}

\usepackage[utf8]{inputenc} 
\usepackage[T1]{fontenc}    
\usepackage{hyperref}       
\usepackage{url}            
\usepackage{booktabs}       
\usepackage{amsfonts}       
\usepackage{nicefrac}       
\usepackage{microtype}      
\usepackage{xcolor}         
\usepackage{paralist}
\usepackage{cleveref}
\usepackage{algorithm,algpseudocode}

\title{Thought Anchors: Which LLM\\Reasoning Steps Matter?}

\author{Paul C. Bogdan\thanks{Equal contribution (author order determined by coinflip).} \\
MATS\\
\And
Uzay Macar\footnotemark[1] \\
MATS\\
\And
Neel Nanda\thanks{Equal senior author contribution (author order determined by coinflip)\\Correspondence: Paul C. Bogdan (\texttt{paulcbogdan@gmail.com}), Uzay Macar (\texttt{uzaymacar@gmail.com})} \\
\And
Arthur Conmy\footnotemark[2] \\
}

\iclrfinalcopy 
\begin{document}

\maketitle

\begin{abstract}

Current frontier large-language models rely on reasoning to achieve state-of-the-art performance. Many existing interpretability are limited in this area, as standard methods have been designed to study single forward passes of a model rather than the multi-token computational steps that unfold during reasoning. We argue that analyzing reasoning traces at the sentence level is a promising approach to understanding reasoning processes. We introduce a black-box method that measures each sentence's counterfactual importance by repeatedly sampling replacement sentences from the model, filtering for semantically different ones, and continuing the chain of thought from that point onwards to quantify the sentence's impact on the distribution of final answers. We discover that certain sentences can have an outsized impact on the trajectory of the reasoning trace and final answer. We term these sentences \textit{thought anchors}. These are generally planning or uncertainty management sentences, and specialized attention heads consistently attend from subsequent sentences to thought anchors. We further show that examining sentence-sentence causal links within a reasoning trace gives insight into a model's behavior. Such information can be used to predict a problem's difficulty and the extent different question domains involve sequential or diffuse reasoning. As a proof-of-concept, we demonstrate that our techniques together provide a practical toolkit for analyzing reasoning models by conducting a detailed case study of how the model solves a difficult math problem, finding that our techniques yield a consistent picture of the reasoning trace's structure. We provide an open-source tool (\href{https://www.thought-anchors.com}{thought-anchors.com}) for visualizing the outputs of our methods on further problems. The convergence across our methods shows the potential of sentence-level analysis for a deeper understanding of reasoning models.

\end{abstract}

\section{Introduction}

Training large language models to reason with chain-of-thought \citep{reynolds2021promptprogramminglanguagemodels, nye2021workscratchpadsintermediatecomputation, wei2023chainofthoughtpromptingelicitsreasoning} has led to significant advances in capabilities \citep{o1openai2024}. The resulting reasoning traces are regularly used in safety research \citep{baker2025monitoringreasoningmodelsmisbehavior, shah2025approachtechnicalagisafety}, but there has been little work adapting interpretability methods to this new paradigm (though see \citep{venhoff2025understanding, goodfire2025r1}). Traditional \textit{mechanistic interpretability} \citep{olah2020zoom, AnthropicMechanisticEssay} methods often focus on a single forward pass of the model: understanding layer-by-layer activations and how these translate into a final output \citep{wang2022interpretabilitywildcircuitindirect, hex2023docstring, hanna2023doesgpt2computegreaterthan}. However, this framework is too fine-grained for autoregressive reasoning models, which consume their own output tokens.

A core step in many interpretability strategies is to decompose the model into smaller parts that can be analyzed independently \citep{lindsey2025biology}. A natural decomposition for chain-of-thought is into individual sentences and how they depend on each other. Interpretations of neural network behavior operate at varying levels of abstraction \citep{geiger1, geiger2}, and sentence-level explanations strike an intermediate abstraction depth. Compared to tokens, sentences are more coherent and often coincide with reasoning steps extracted by an LLM  \citep{venhoff2025understanding, arcuschin}. Compared to paragraphs, sentences are less likely to conflate reasoning steps and may serve as an effective target for linking different steps.

\begin{figure}
  \centering
  \includegraphics[width=1\linewidth]{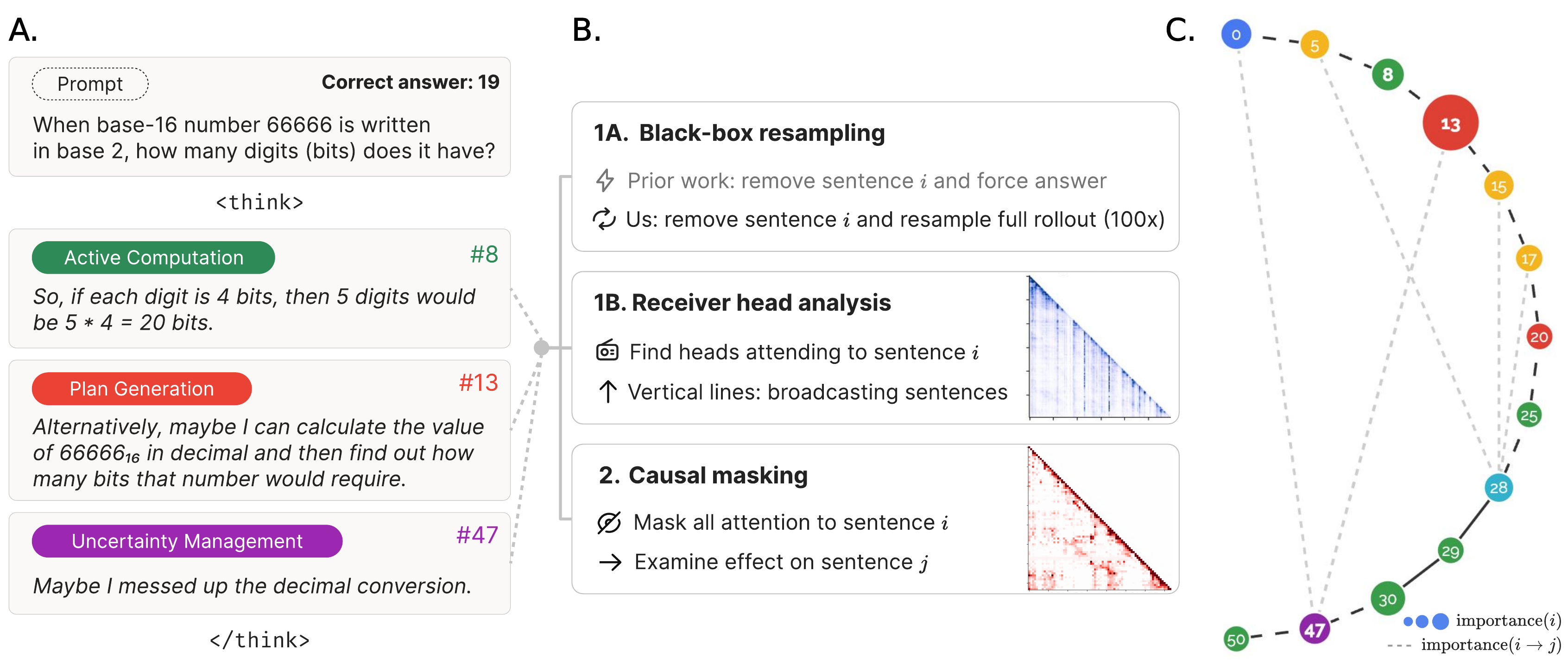}
  \caption{Summary of our methods for principled attribution to important sentences in reasoning traces. \textbf{A.} An example reasoning trace with sentences labeled per our taxonomy. \textbf{B.} Our proposed methods are: black-box resampling, receiver heads, and attention suppression. \textbf{C.} A directed acyclic graph among sentences prepared by one of our techniques, made available open source.} 
  \label{figure:study_summary}
\end{figure}

Prior work has established that CoT contains reasoning steps performing distinct functions. Backtracking sentences (e.g., \textit{``Wait…''}) cause the model to revisit earlier conclusions, which boosts answer accuracy \citep{muennighoff2025s1simpletesttimescaling}. Other research has distinguished sentences based on whether they retrieve new information or execute deduction with existing information \citep{venhoff2025understanding}. Hence, reasoning may follow an overarching structure, where computational goals are generated, revised, and pursued. Yet, approaches for mapping this high-level structure are limited.

We propose that reasoning traces can be understood through \textbf{thought anchors}: critical reasoning steps that guide the trajectory of reasoning. We provide evidence for this type of anchoring based on black-box evidence from resampling and white-box evidence based on attention patterns. By measuring the causal dependencies between sentences via a masking approach, we further show how a CoTs wider computational structure can be interpreted. These measures go beyond just reading a CoT's text, providing a principled foundation for interpretability that sidesteps disputes about the ``faithfulness'' of CoT text \citep{turpin2023languagemodelsdontsay,korbak2025chain}. 

\Cref{secBlackBoxFramework} and \Cref{sentence-systematic} introduce a black-box method for measuring the counterfactual impact of a sentence on the model's final answer and future sentences. We repeatedly resample reasoning traces from the start of each sentence. Based on resampling data, we can quantify the counterfactual impact of each sentence on the likelihood of any final answer. Additionally, we can distinguish planning sentences that initiate computations leading to some answer from sentences performing computations necessary for the answer but which are predetermined. \Cref{secAttentionAggregation}, adds a white-box method for evaluating importance based on the sentences most attended. Our analyses reveal ``receiver'' heads that narrow attention toward particular past ``broadcasting'' sentences. This provides a mechanistic measure of importance, whose findings converge with our resampling technique.

\Cref{sec:attention-suppression} and \Cref{systematic-masking} present a method mapping the wider structure of the reasoning the causal dependencies between pairs of CoT sentences. For each sentence in a trace, we intervene by masking all attention to it from subsequent tokens or simply removing the sentence entirely. We then measure the effect on subsequent token logits (KL divergence) compared to those generated without masking. Averaging token effects by sentence, this strategy measures each sentence's direct causal effect on each subsequent sentence.

Applying these techniques, our work suggests that analyzing reasoning through sentence-level units introduces new domains through which reasoning models can be understood. Our work also opens the door to more precise debugging of reasoning failures, identification of sources of unreliability, and the development of techniques to enhance the reliability of reasoning models.


\section{Quantifying sentence importance}
\label{secBlackBoxFramework}

Some sentences matter more than others, but which ones matter most depends on how we define and measure importance. We frame sentence-level importance as a question of counterfactual influence: how does including or excluding a sentence affect subsequent steps and the model's final output?

\subsection{Model and dataset}
\label{subsecModelAndData}

Our analyses of sentence importance are based on the DeepSeek R1-Distill Qwen-14B model (48 layers) \citep{deepseekai2025deepseekr1incentivizingreasoningcapability}. We used a temperature of $0.6$ and a top-p value of $0.95$. We focus on the MATH dataset \citep{hendrycks2021measuringmathematicalproblemsolving}. Our analysis hinges on variability in final responses, so we target 20 challenging but doable questions that are correctly solved 25-75\% of the time, identified by testing on 1,000 problems 10 times each.
For each selected problem, we generated one correct and one incorrect reasoning trace, producing 40 responses. The average response is 144.2 sentences (95\% CI: [116.7, 171.8]) and 4208 tokens (95\% CI: [3479, 4937]). We focus only on sentences before the model has converged on an answer (i.e., after which it gives the same response in >98\% of resamples). In  \Cref{appendix-other-models}, we provide results from applying our techniques to the R1-Distill-Llama-8B model.

\subsection{Forced answer importance}

Earlier work has measured sentence importance by forcing a model to answer before completing its reasoning trace \citep{lanham2023measuringfaithfulnesschainofthoughtreasoning}. We compared our approach to this existing technique: For each sentence in a CoT, we interrupt the model and append text, inducing a final output (\texttt{``Therefore, the final answer is \textbackslash boxed\{}''). This is done 100 times at each sentence position.

\subsection{Importance via resampling}
\label{subsecCounterfactualImportance}

A limitation of the forced-answer approach is that a sentence $S$ may be necessary for some final answer but is consistently produced by the LLM late in the reasoning trace (e.g., a reliable arithmetic statement). This means that forced answer accuracy will be low for all sentences before $S$, precluding earlier step importance from being assessed. 

Our approach evaluates importance by examining how a sentence may guide downstream sentences. Consider a rollout consisting of sentences $S_1, S_2, \dots , S_i, \dots, S_M$ and a final answer $A$. We can use resampling to capture the extent sentence $S$ influences $A$. Specifically, for a given sentence $S_i$, we generate a distribution over final answers by generating 100 rollouts both without sentence $S_{i}$ (the base condition, with rollouts of the form $S_1, S_2, \dots, S_{i-1}, T_i, \dots, T_N, A'_{S_i}$), and another distribution with sentence $S_{i}$ (the intervention condition, with rollouts of the form $S_1, S_2, \dots, S_{i-1}, S_i, \dots, S_M, A_{S_i}$). To assess the utility of resampling, we first conducted a brief case study.

\subsection{Case study}
\label{case-resampling}

\begin{figure}
  \centering
  \includegraphics[width=0.93\linewidth]{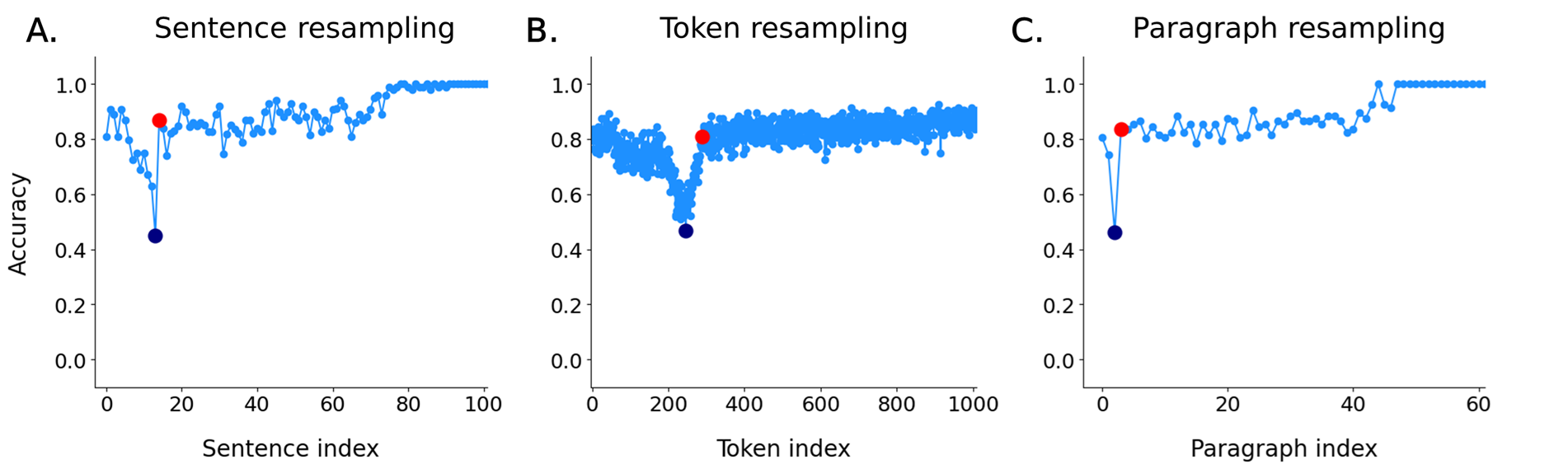}
  \caption{Accuracy over 100 rollouts at each (\textbf{left}) token, (\textbf{middle}) sentence, or (\textbf{right}) paragraph. Navy and red circles border the most importance sentence (Sentence 13) and are plotted in each graph as a reference. For the token graph, resampling was only done on the first 1,000 tokens of the CoT.}
  \label{figure:resampling_examples}
\end{figure}

We first investigate the efficacy of our sentence importance technique by applying it to one problem: \textit{``When the base-16 number $66666_{16}$ is written in base 2, how many base-2 digits (bits) does it have?''} (MATH Problem 4682; see \Cref{appendix-case-transcript} for the CoT transcript). The resampling data shows that from sentences 6-12, expected accuracy steadily declines, but sentence 13 causes accuracy to drastically increase (indicated by the navy and red circles in Figure~\ref{figure:resampling_examples}A). 

The large accuracy fluctuation motivates inspection of this part of the CoT. The model initially considers that $66666_{16}$ contains five base-16 digits, and any base-16 digit can be represented with four base-2 digits. Thus, the model considers the answer: 20 bits. However, this overlooks that $6_{16}$ is $110_{2}$ rather than $0110_{2}$ due to the leading zero. Interestingly, Sentence 12 mentions \textit{``checking if there’s any leading zero that might affect the bit count,''} yet Sentence 12 lowers the expected accuracy. The uplift comes from Sentence 13, where the model decides to \textit{``calculate the value of $66666_{16}$ in decimal''} (see resample alternatives in \Cref{appendix-sentence-13-alternatives}). Downstream reasoning computes the decimal value of $66666_{16}$ and converts it to binary to arrive at the correct answer: 19 bits. The pivotal role of Sentence 13 is instead missed if one examines forced-accuracy importance (Figure~\ref{figure:resampling_examples}A). This case study provides initial evidence that our resampling-based strategy identifies key moments in a CoT, where impactful plans are set or modified.  

Further analyses show the efficacy of specifically examining sentences. The sentence-level resampling data mirrors the patterns seen resampling tokens but at a fraction of the cost (Figure~\ref{figure:resampling_examples}B). Resampling paragraphs leads to meaningfully less resolution (Figure~\ref{figure:resampling_examples}C). Although future work may benefit from defining reasoning steps at adaptive scales, the current results suggest that considering sentences provides high resolution while being cheaper than tokens.


\section{Consistent patterns in sentence importance}
\label{sentence-systematic}

\subsection{Sentence taxonomy}
\label{subsecSentenceTaxonomy}

To more systematically test whether reasoning is characterized by key sentences with outsized impacts, we organized sentences into different categories and measured their causal impacts. We adopted the framework by \cite{venhoff2025understanding}, which describes distinct reasoning functions within a reasoning trace. We define eight categories (see examples and frequencies in \Cref{appendix-taxonomy}): 

\begin{compactenum}
    \item \textbf{Problem Setup:} Parsing or rephrasing the problem
    \item \textbf{Plan Generation:} Stating or deciding on a plan of action, meta-reasoning
    \item \textbf{Fact Retrieval:} Recalling facts, formulas, problem details without computation
    \item \textbf{Active Computation:} Algebra, calculations, or other manipulations toward the answer
    \item \textbf{Uncertainty Management:} Expressing confusion, re-evaluating, including backtracking
    \item \textbf{Result Consolidation:} Aggregating intermediate results, summarizing, or preparing
    \item \textbf{Self Checking:} Verifying previous steps, checking calculations, and re-confirmations
    \item \textbf{Final Answer Emission:} Explicitly stating the final answer
\end{compactenum}

Each sentence in the analyzed response is assigned to one of these categories using an LLM-based auto-labeling approach (detailed in \Cref{appendix-for-prompt}). Categories that rarely appear are omitted from the figures below. Residual-stream probes accurately distinguish categories (see \Cref{appendix-probing}).


\subsection{Counterfactual importance}

We additionally formalize our approach to quantifying importance in a manner that can be applied to any problem, including ones with any number of possible outcomes. We present two measures:

\begin{compactenum}
    \item \textbf{Resampling importance.} We can compute the KL Divergence between the final answer distributions in the two conditions, i.e., $\text{importance}_{\text{r}} := D_{\text{KL}}[ p(A'_{S_i}) || p(A_{S_i}) ]$, providing a scalar measure of how much sentence $S_i$ changes the answer. We call this \emph{resampling importance}. We include $\epsilon = 10^{-9}$ to avoid division by zero, but the below conclusions remain consistent if instead performed using additive smoothing ($\alpha = 0.5$ or $1.0$). 
    \item \textbf{Counterfactual importance.} The problem with resampling importance is that if $T_i$ is identical or similar to $S_i$ then we do not get much information about whether $S_i$ is important or not. Therefore, we write $S \not\approx T$ if two sentences $S$ and $T$ are dissimilar, defined as having embeddings with a cosine similarity less than the median value across all sentence pairs in our dataset (see \Cref{appEmbeddings} for details). Therefore, we can define \emph{counterfactual importance} by conditioning on $T_i \not\approx S_i$; i.e., $\text{importance} := D_{\text{KL}}[ p(A'_{S_i} | T_i\not\approx S_i) || p(A_{S_i}) ]$.
\end{compactenum}


Because we resample all steps after a given sentence $S_i$, we avoid the aforementioned limitation of forced-answering. We also provide empirical evidence that the principled \textbf{counterfactual importance} definition in 1-3 above is useful, by comparing it to the resampling importance in \Cref{appendix-resample-methods}. As a comparison, we also evaluate \emph{forced answer importance} based on this KL divergence strategy.

\subsection{Results}

\begin{figure}
  \centering
  \includegraphics[width=1.0\linewidth]{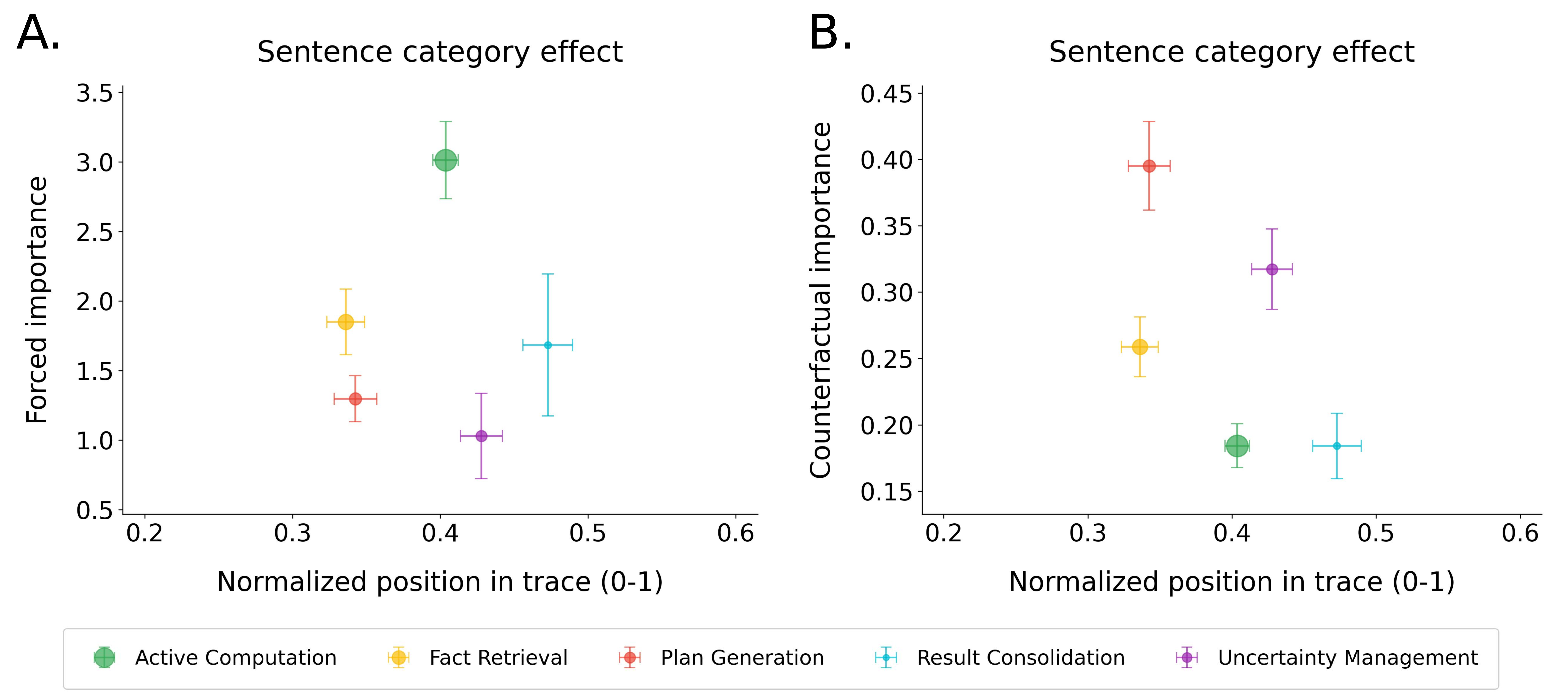}
  \caption{The mean of each sentence category for (\textbf{A}) forced-answer importance and (\textbf{B}) counterfactual importance, per the resampling method, plotted against the sentence category's mean position in the reasoning trace. Only the 5 most common sentence types are shown (see \Cref{appendix-resampling}).}
  \label{figure:sentence_category_importances}
\end{figure}

\textit{Plan generation} and \textit{uncertainty management} (e.g., backtracking) sentences consistently show higher counterfactual importance than other categories like \textit{fact retrieval} or \textit{active computation} (see Figure~\ref{figure:sentence_category_importances}B). This supports the view that high-level organizational sentences anchor, organize, and steer the reasoning trajectory. These findings deviate from the analysis of forced answer importance, which instead implicates \textit{active computation} as producing the greatest distributional shifts (Figure~\ref{figure:sentence_category_importances}A). The forced-answer approach entirely neglects the importance of planning that influences other sentences, which we argue is more meaningful for understanding the trajectory of a reasoning trace.

\section{The mechanistic roots of importance} 
\label{secAttentionAggregation}


We hypothesize that important sentences may receive heightened attention from downstream sentences. Although attention weights do not necessarily imply causal links (see Section~\ref{sec:attention-suppression}), heightened attention is a plausible mechanism by which important sentences influence subsequent sentences. We conjecture further that a high focus on important sentences may be driven by specific attention heads, and by tracking such heads, we may pinpoint key sentences.

\begin{figure}
  \centering
  \includegraphics[width=0.9\linewidth]{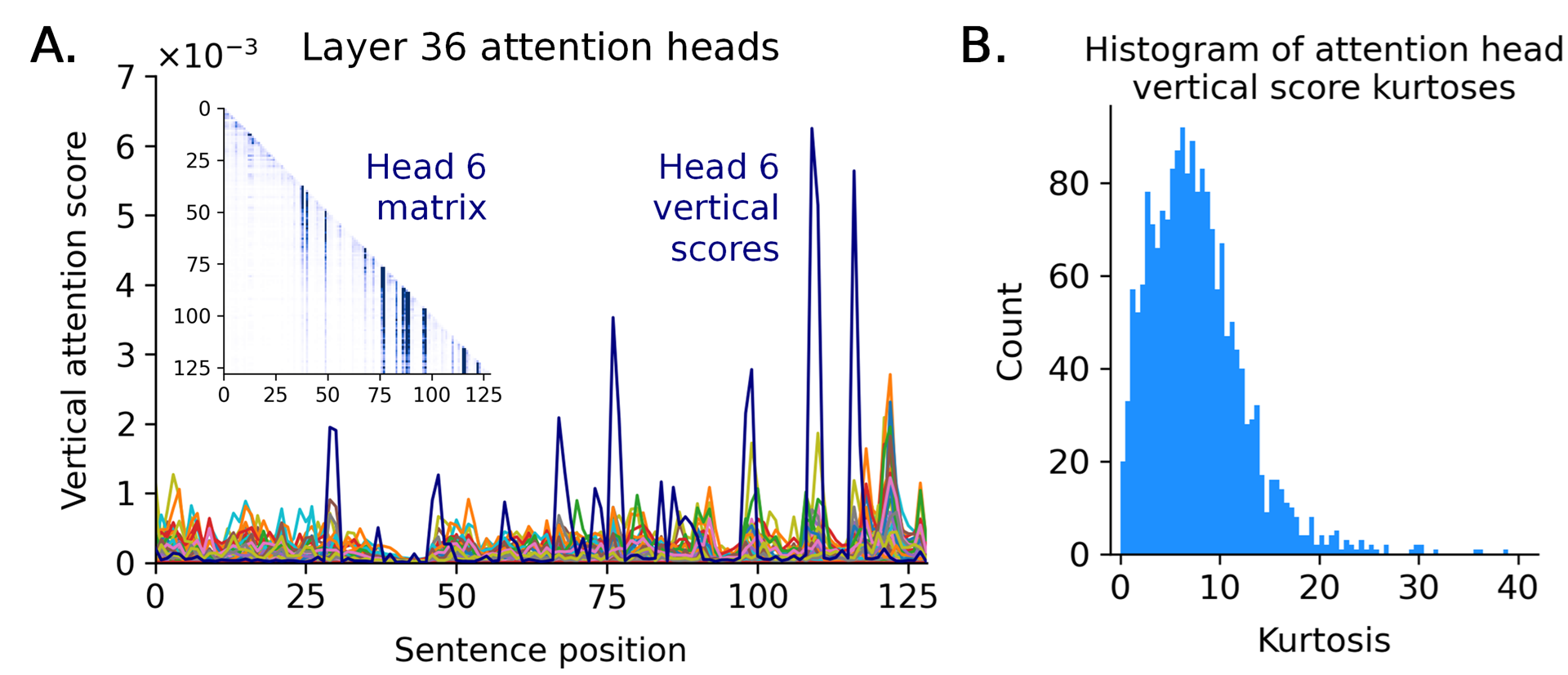}
  \caption{\textbf{A}. Lines show the vertical attention scores for each sentence by the 40 different heads in layer 36. Head 6 has been highlighted as a receiver head, and its corresponding attention weight matrix is shown for reference. Its prominent spikes cause the distribution to have a high kurtosis.  \textbf{B}. Histogram of these kurtosis values across all attention heads, median across all reasoning traces.}
  \label{figure:rec_dist}
\end{figure}

We assessed the degree different heads narrow attention toward particular sentences. First, for each reasoning trace, we averaged each attention head's token-token attention weight matrix to form a sentence-sentence matrix, where each element is the mean across all pairs of tokens between two sentences. Based on each attention matrix, we computed the mean of its columns below the diagonal to measure the extent each sentence receives attention from all downstream sentences; averaged only among pairs at least four sentences apart to focus on distant connections. This generates a distribution for each head (e.g., Figure~\ref{figure:rec_dist}A), and the extent each head narrows attention toward specific sentences can be quantified as its distribution's kurtosis (computed for each reasoning trace, then averaged across traces). Plotting each head's kurtosis reveals that some attention heads strongly narrow attention toward specific, possibly important, sentences in the reasoning trace (Figure~\ref{figure:rec_dist}B).

\subsection{The identification of receiver heads}

We refer to attention heads that narrow attention toward specific sentences as \textit{``receiver heads''}. These heads are more common in later layers (\Cref{appendix-receiver-heads}). To formally assess the existence of receiver heads, we tested whether some attention heads consistently operate in this role by measuring the split-half reliability of heads' kurtosis scores. We found a strong head-by-head correlation ($r$ = .84) between kurtosis scores computed for half of the problems with kurtosis scores for the other half of problems. Thus, some attention heads consistently operate as receiver heads, albeit with some heterogeneity across responses in which heads narrow attention most.

Receiver heads usually direct attention toward the same sentences. Among the 16 heads with the highest kurtoses, we computed the sentence-by-sentence correlation between the vertical-attention scores for each pair of heads; calculated separately for each reasoning trace, then averaged. This produced an large correlation (mean $r$ = .56). Thus, receiver heads generally attend the same sentences (for reference, the average correlation among any heads is $r$ = .35). This convergence across receiver heads is consistent with the existence of sentence importance, which these heads identify.

Attentional narrowing toward particular sentences may be a feature specifically of reasoning models that enhances their performance. Comparing R1-Distill-Qwen-14B (reasoning) and Qwen-14B (base) suggests that the reasoning model's receiver heads will narrow attention toward singular sentences to a greater degree (\Cref{rec-head-difs}). Furthermore, ablating receiver heads leads to a greater reduction in accuracy than ablating self-attention heads at random (\Cref{appendix-ablation}). Altogether, these findings are consistent with receiver heads and thought-anchor sentences playing special roles in reasoning.

\subsection{Links to counterfactual importance and sentence types}

\begin{figure}[t]
  \centering
  \includegraphics[width=0.7\linewidth]{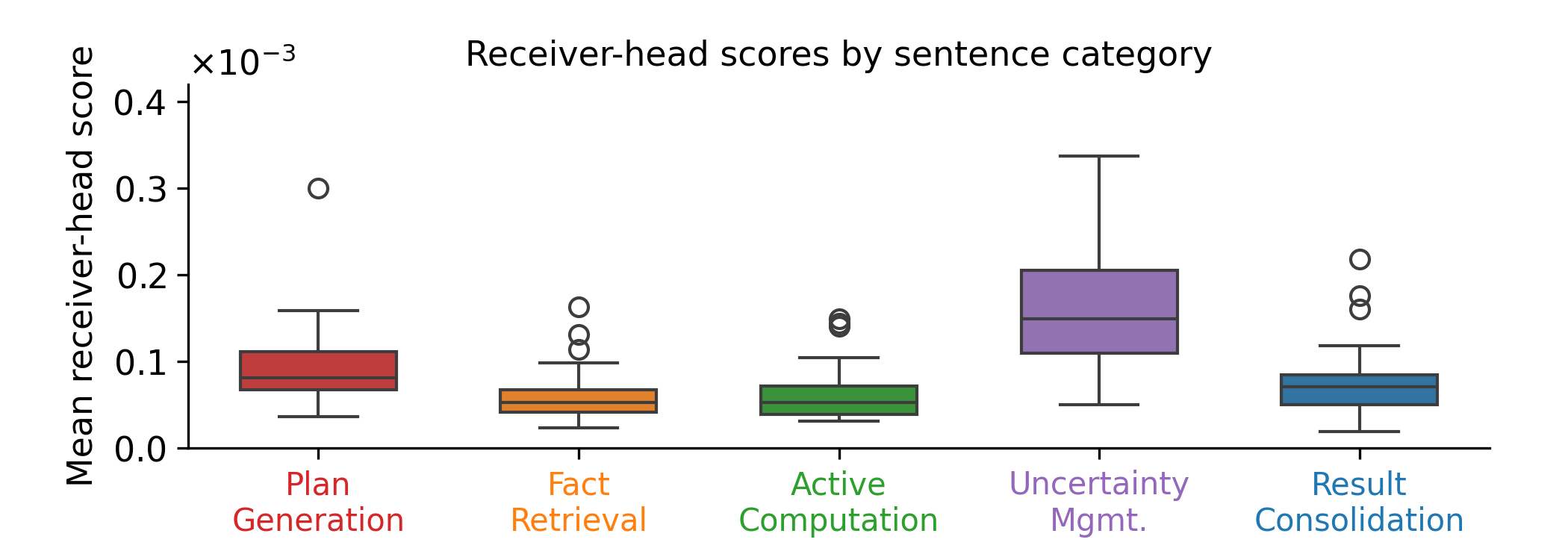}
  \caption{The boxplot shows the average top-32 receiver-head score for each sentence type. The boxes correspond to the interquartile range across different reasoning traces.}
  \label{figure:rec_tax}
\end{figure}

\textit{Plan generation} and \textit{uncertainty management} sentences consistently receive the most attention via receiver heads (Figure~\ref{figure:rec_tax}), whereas \textit{active computation} sentences receive relatively minimal attention ($t$s > 4.0, $p$s < .001 per paired t-tests comparing the mean receiver-head score for the former two versus the later two categories). These findings demonstrate a parallel between the receiver head findings here and the earlier results on the sentence types yielding the highest counterfactual importance.


\section{Sentence-sentence causal links}
\label{sec:attention-suppression}

\subsection{Approach}

We next focused on how key sentences influence specific subsequent sentences. We used a sentence-masking strategy,
suppressing all attention (all heads) to a given sentence and examining how this impacts future sentences, measured using the KL divergence between token logits with or without masking. The overall effect on a future sentence is the average of its token log-KL divergences. We normalize this score by subtracting the latter sentence's average causal effect from all prior sentences. Suppressing attention is mostly equivalent to omitting a sentence from a CoT, only differing in positional embeddings; computing the causal graph as so can be done with LLM APIs that return logits without exposing attention. \Cref{appendix-graph-pseudocode} provides pseudocode for generating the causal graph.

Our masking approach assumes (i) token logits capture a sentence's semantic content and (ii) masking sentences does not problematically induce out-of-distribution behavior. We evaluated these assumptions by correlating the sentence-sentence scores with those from an alternative strategy based on our counterfactual resampling method, which assesses how resampling $S_i$ with $T_i$ ($S \not\approx T$) influences the likelihood of $S_j$ appearing. This measure positively correlates with the scores from the masking- \& logits-based strategy (detailed further in \cref{appendix-sentence-sentence-counterfactual}), suggesting that logits indeed track semantics despite simulating out-of-distribution behavior. We continue with the sentence-masking approach because it requires $\sim$100x less compute than resampling, increasing scalability.

\subsection{Case study}
\label{secCaseStudy}

\begin{figure}[h]
\centering
  \includegraphics[width=0.7\linewidth]{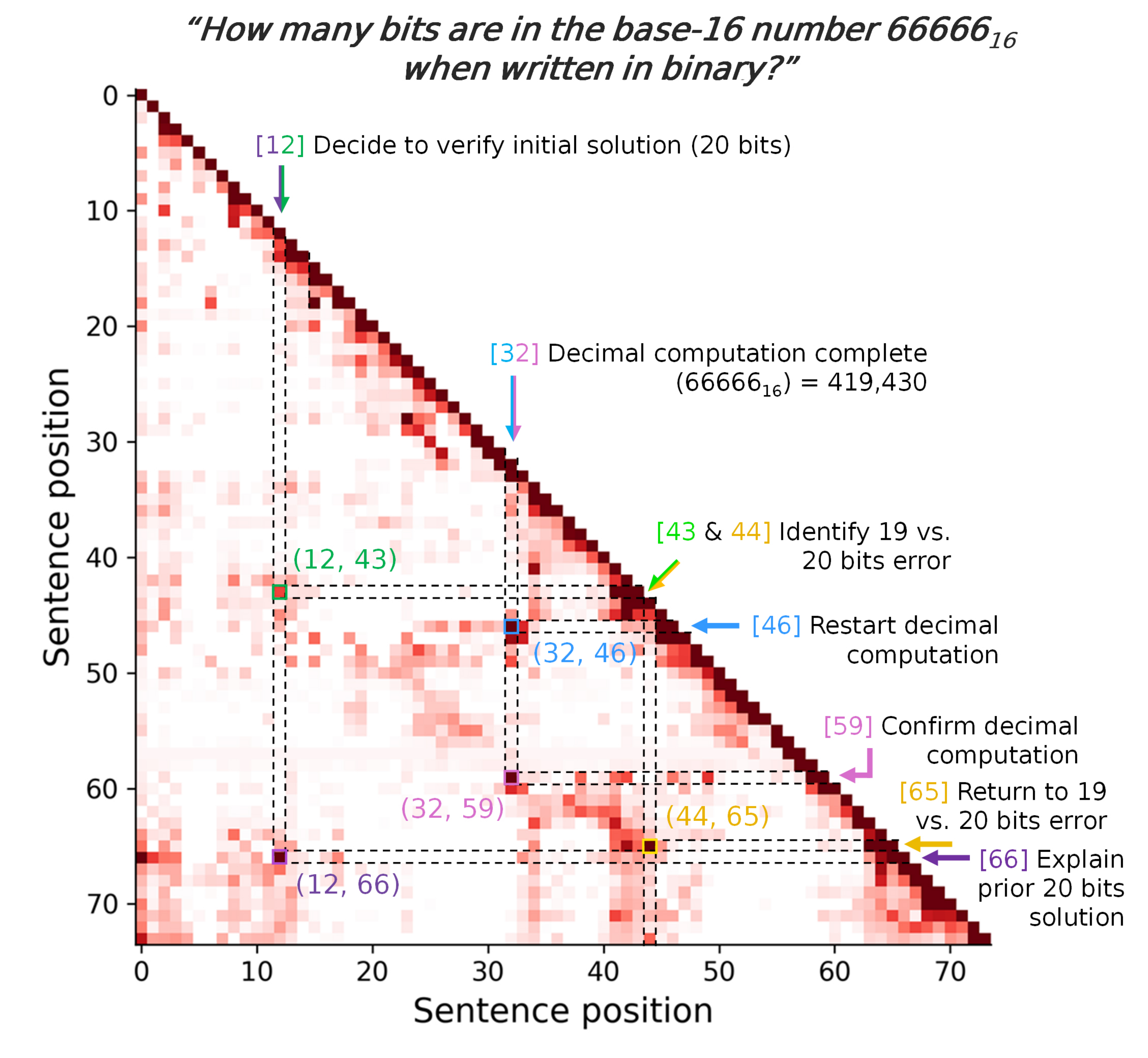}
  \caption{For the correct-answer CoT of Problem \#4682, the matrix shows the effect of masking one sentence (x-axis) on a future sentence's logits (y-axis). Darker colors indicate higher values.}
  \label{figure:case-graph}
\end{figure}

We begin with a small-scale investigation to provide intuition for our sentence-sentence measure and motivate more systematic tests. We continue our initial case study (\Cref{case-resampling}), but here, we focus 
on three local maxima in the sentence-masking graph (Figure~\ref{figure:case-graph}), which align closely with the sentences implicated as important by receiver-heads (see further details on the case study in \Cref{appendix-case-depth}):


\begin{compactitem}
\item\textbf{(Sentences: 12 → 43)} After suggesting the answer ``20 bits'', the model decides to begin verifying it (Sentence 12). Verification leads to a different solution, ``19 bits'' (Sentence 43). Between these key sentences, most of the intermediate text is performing arithmetic. 
\item\textbf{(Sentences: 44 → 65)} Noticing the discrepancy (Sentence 44), the model decides to check its calculations. It finds that they are correct, and the discrepancy remains (Sentence 65). 
\item\textbf{(Sentences: 12 → 66)} The model realizes that its initial suspicion about leading zeroes (Sentence 12) is justified and states that this is the reason for the discrepancy (Sentence 66). 
\end{compactitem}

These connections point to an interpretable scaffold reflecting computations on the pursuit of intermediate results, the execution of self-correction subroutines, and the synthesis of prior statements.

\subsection{Open source interface}

We released an open source interface 
(\href{https://www.thought-anchors.com}{thought-anchors.com}) 
for visualizing reasoning traces and comparing alternative rollouts. We show our proof-of-concept interface in Figure~\ref{figure:study_summary}C, where important sentences are represented by larger nodes and causal connections between sentences are shown with dashed gray lines. The tool aims to benefit interpretability and unwanted behavior debugging.

\section{Systematic differences in sentence-sentence relationships}
\label{systematic-masking}

The case study demonstrates that pivotal moments within a CoT -- e.g., where a conclusion is posed, where a decision is made about the next steps to answering a question, or where a discrepancy with earlier information is identified -- are sensitive to historic information in the CoT and/or exert impacts far downstream. We next investigated how causal graphs may shed light on questions about reasoning in LLMs generally. Specifically, we ask: How many examining sentence-sentence links shed light on model confidence during reasoning? Relatedly, why do some problem domains like mathematics display stronger uplift in reasoning compared to non-reasoning models? 

We hypothesize that strong linkages between nearby sentences reflect a coherent logical flow and well-formed plan, so each sentence causes the next, whereas distant linkages reflect uncertainty and backtracking. Despite occasional long-range connections, we further hypothesize that successful mathematical CoTs are specifically characterized by strong close links between sequential sentences, whereby planning statements sharply structure the CoT. Domains related to mathematics may uniquely lend themselves to such firmly-structured reasoning, whereas CoTs for other topics (e.g., history or biology) may solve problems principally by scanning a wide latent space in a less structured fashion. 


\subsection{Methods}

We pivoted to analyzing MMLU problems \citep{hendrycks2020measuring}, so that we could contrast problem domains. We also switched to Qwen3-30b-a3b, so that we could leverage a serverless LLM provider that outputs token logits, which allowed scaling up our analysis to thousands of CoTs. We ran Qwen3-30b-a3b in non-reasoning mode on all 15,638 MMLU questions to identify challenging problems where non-reasoning accuracy is under 50\% (per answer logits). This corresponds to 3,651 problems, and for 2,492 of these questions, the model answers correctly when using reasoning at least once across ten passes. We computed each correct CoT's causal graph ($M_{sentences}$ = 90.1).

We compared graphs on the strength of their causal links at different distances between sentences. We specifically computed the mean attention-suppression effect at distance $k$ for each graph ($m \times m$ sentences) for all $k \le \frac{m}{2}$. This corresponds to the mean of a matrix's $k$-th subdiagonal. We consider subdiagonals only up to $\frac{m}{2}$ to reduce noise by ensuring that the mean is computed among an adequate number of elements (e.g., the $m$-th subdiagonal would be just the single bottom-leftmost element).

\subsection{Results}

\begin{figure}[h]
  \centering
  \includegraphics[width=1.0\linewidth]{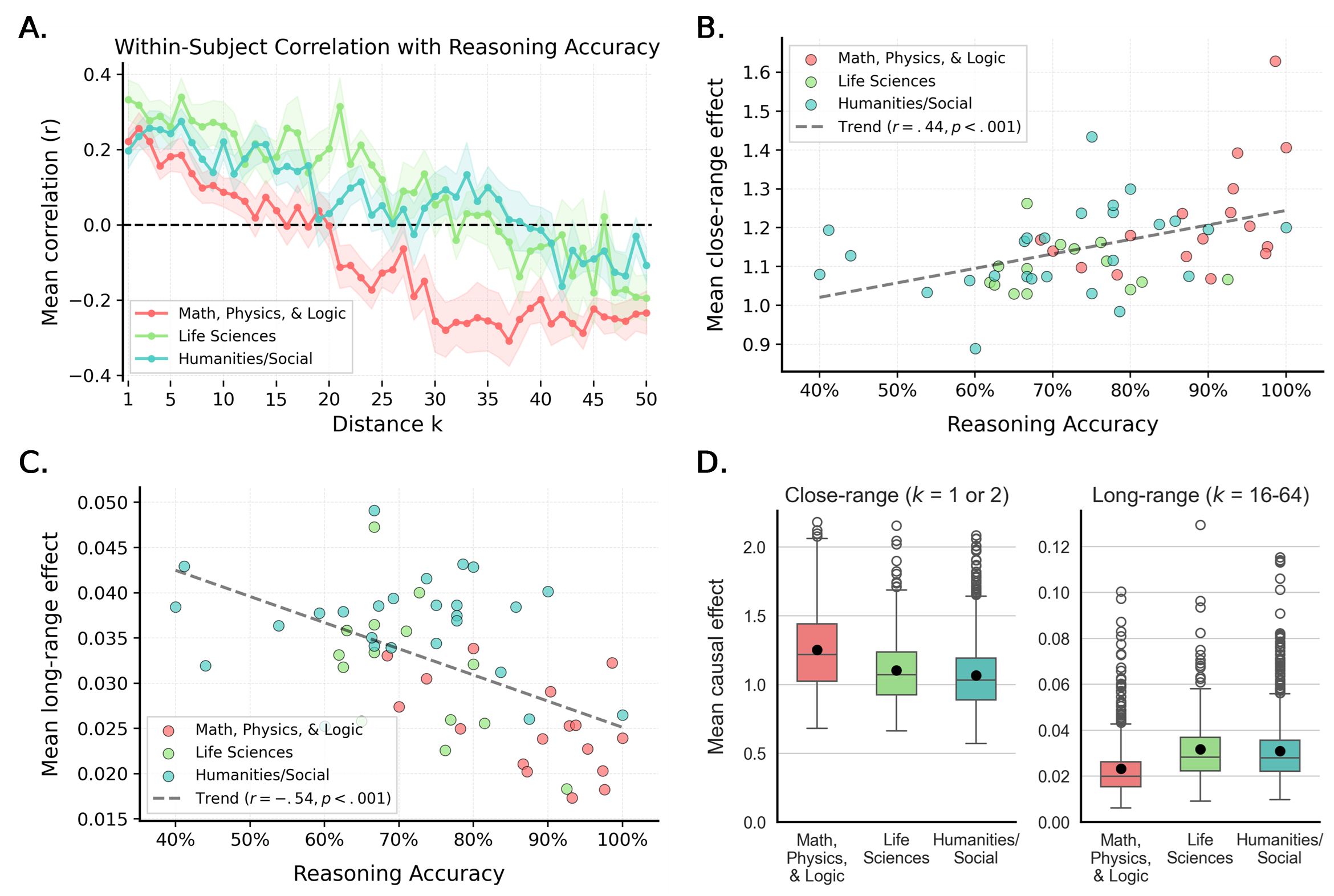}
  \caption{\textbf{A.} For each distance $k$, we computed the correlation between a question's average $k$-distance causal effect in one CoT and the question's mean reasoning accuracy across ten CoTs. \textbf{B. \& C.} Scatterplot shows each subject's average close-range ($k$ = 1-2) and long-range ($k$ = 16-64) was plotted against its average reasoning accuracy. \textbf{D.} 
  Box-plots showing the spread of average close-range and long-range causal effects for different question domains; each point represents one CoT, and black circles represent means.}
  \label{figure:mmlu}
\end{figure}

The distance of causal effects tracks question difficulty. Computing correlations within-subject, we find that questions with high average accuracy elicit CoTs with stronger close-range links and weaker long-range links (Figure~\ref{figure:mmlu}A). In addition, subjects where average accuracy is high overall tend to produce CoTs with stronger close links ($r = .44, p < .001$; Figure~\ref{figure:mmlu}B) and weaker long links ($r = -.54, p < .001$ Figure~\ref{figure:mmlu}C). The strongest levels of accuracy were seen in problems requiring mathematical thinking (e.g., mathematics \& physics). As hypothesized, these areas also yielded CoTs with stronger close-range connections and weaker long-range connections (two-sample t-test $|t|s > 10, ps < .001$; Figure~\ref{figure:mmlu}D). Although these analyses do not model \textit{plan generation} and \textit{uncertainty management} (thought anchors) sentences directly, the present findings speak to their potential structural roles and overall shed light on the nature of successful reasoning.




\section{Related work}

\paragraph{Reasoning advances and unfaithfulness in LLMs.} CoT reasoning, optimized using reinforcement learning, has driven major capabilities improvements in large language models \citep{wei2023chainofthoughtpromptingelicitsreasoning, nye2021workscratchpadsintermediatecomputation, reynolds2021promptprogramminglanguagemodels}.
This reasoning paradigm introduces novel safety challenges. Experiments inducing unfaithful reasoning have led some to raise concerns about the interpretability of CoT text \citep{lanham2023measuring, chen2025reasoningmodelsdontsay}, although others have argued that CoT text generally is a meaningful representation, particularly for difficult tasks \citep{korbak2025chain}. By showing how sentence types, categorized based on their text, differ in their resampling and receiver-head importance, our findings endorse the meaningfulness and interpretability of CoT text. 

\textbf{Importance of individual steps.} A variety of prior techniques that can be used for CoT interpretability have been developed, and these likewise have suggested that a subset of steps disproportionately drive the final answer -- e.g., findings based on 
Shapley values \citep{gao2023shapleycot}, ROSCOE metrics \citep{golovneva2023roscoesuitemetricsscoring}, gradient-based scores \citep{wu2023analyzingchainofthoughtpromptinglarge}, and resampling at fork tokens \citep{bigelow2024forkingpathsneuraltext}. Complementing these, our approach provides a more principled framework for understanding how reasoning traces are constructed around key sentences.


\section{Discussion and Limitations}
\label{secDiscussion}

This work presents initial steps towards a principled decomposition of reasoning traces with a focus on identifying thought anchors: sentences with outsized importance on the model's final response, specific future sentences, and downstream reasoning trajectory. We have also begun unpacking the attentional mechanisms associated with these important sentences. We expect that understanding thought anchors will be critical for interpreting reasoning models and ensuring their safety.

While some research raises valid concerns that CoT text can be unfaithful to the model's underlying computation \citep{lanham2023measuring, chen2025reasoningmodelsdontsay}, our results show CoT text is mechanistically relevant and interpretable. For example, sentences categorized as \texttt{plan generation} and \texttt{uncertainty management} consistently exhibit higher counterfactual importance in our resampling analyses and receive more focused attention from receiver heads. This demonstrates a link between what a sentence says and its functional role in the computation, and this type of correspondence supports arguments on the value of CoT legibility \citep{korbak2025chain}.

A primary limitation of our resampling approach is its computational cost. In this work, we resampled 100 times per sentence to achieve fairly precise estimates (in terms of final-answer accuracy, 95\% CI corresponds to at worst $\pm10\%$), which was sufficient to identify pivotal moments in the case study (Figure~\ref{figure:resampling_examples}A). However, for analyses focusing on aggregate patterns across many CoTs, fewer resamples for any one CoT may suffice. Additionally, future work could develop adaptive resampling strategies that allocate more computational budget to potentially pivotal moments in the trace, maximizing precision while minimizing cost.

We view this as preliminary work. Our analyses require refinement to fully grapple with how downstream sentences may be overdetermined by different trajectories in a reasoning trace or independent sufficient causes. Additionally, we do not formally examine the role of error correction. Our receiver-head analyses are confounded by a sentence's position in the reasoning trace (see \Cref{appendix-position-rec}). Further, our attention-suppression work is limited because it effectively requires the model to process out-of-distribution information. 

Despite these limitations, we believe that we have demonstrated that our metrics are an advance on prior work, interrupting models and forcing final answers. The surprising degree of shared structure we have found across our three methods illustrates the potential value of future work in this area and points to the possibility of more powerful interpretability techniques to come.

\section{Reproducibility Statement}

To ensure the reproducibility of our work, we provide comprehensive implementation details, code, and experimental specifications. Our code is publicly available at \url{https://github.com/interp-reasoning/thought-anchors/}
, which includes all scripts for black-box resampling, receiver head analysis, and attention suppression experiments. We also provide an interactive visualization tool at \url{https://www.thought-anchors.com} 
for exploring reasoning traces and sentence-level causal dependencies. The complete prompt used for sentence taxonomy labeling is provided in \Cref{appendix-for-prompt}, including detailed instructions for function tags and dependency annotations. Our experimental setup uses DeepSeek R1-Distill-Qwen-14B (48 layers) with temperature 0.6 and top-p 0.95, tested on the MATH dataset \citep{hendrycks2020measuring} focusing on problems with 25-75\% solution rates. We specify exact hyperparameters including 100 rollouts per sentence for counterfactual resampling, cosine similarity threshold of 0.8 (median value) using \texttt{all-MiniLM-L6-v2} embeddings (\Cref{appEmbeddings}), and identification of receiver heads via kurtosis scores of attention distributions. The sentence-sentence causal masking methodology is fully detailed in \Cref{sec:attention-suppression}, with validation through correlation with resampling-based measures (\Cref{appendix-sentence-sentence-counterfactual}). For MMLU experiments in \Cref{systematic-masking}, we used Qwen3-30b-a3b on 2,492 problems where non-reasoning accuracy is below 50\%, computing causal graphs for correct CoTs. Additional reproducibility details include: full case study transcript (\Cref{appendix-case-study}), sentence category distributions (\Cref{appendix-taxonomy}), receiver head ablation procedures with 128/256/512 heads (\Cref{appendix-ablation}), and cross-model validation on R1-Distill-Llama-8B (\Cref{appendix-other-models}). All models used are publicly available, and we provide pseudocode for the sentence-to-sentence importance calculation in \Cref{appendix-sentence-sentence-counterfactual}.

\section*{Acknowledgments}

This work was conducted as part of the ML Alignment \& Theory Scholars (MATS) Program. We would like to thank Iván Arcuschin, Constantin Venhoff, and Samuel Marks for helpful discussions and feedback. We particularly thank Stefan Heimersheim for his valuable feedback and insightful suggestions, including ideas for experimental approaches that helped strengthen our analysis and contributed to the clarity of our presentation. We also thank members of Neel Nanda's MATS stream for engaging in brain-storming sessions, thoughtful questions during our presentations, and ongoing discussions that helped shape our approach.

\section*{Author Contributions}

Both first authors, Paul C. Bogdan and Uzay Macar, contributed to the research, engineering, and writing of the paper. Neel Nanda was the main supervisor and provided feedback and guidance throughout the project. Arthur Conmy proposed the black-box resampling method for measuring the importance of sentences and also provided feedback and guidance throughout the project.

\bibliography{iclr2026_conference}
\bibliographystyle{iclr2026_conference}

\newpage
\appendix

\section{Evaluating importance (KL) while smoothing}
\label{appendix-smoothing}

The identified link between sentence's category and its forced-answer or counterfactual importances were also measured while smoothing the final-answer distribution associated with each sentence. Smoothing was performed when computing the KL divergence between the two distribution and constitutes replacing the $\epsilon = 10^{-9}$ term (originally used to avoid division by zero) with $\alpha = 1.0$ (Laplace smoothing) or $\alpha = 0.5$ (smoothing with Jeffrey's prior). 

Let $p(A'_{S_i})$ and $p(A_{S_i})$ be the empirical distributions over a set of $K$ possible final answers, $\mathcal{A}$, derived from $N$ rollouts (e.g., $N=100$). Let $C'_{S_i}(a)$ and $C_{S_i}(a)$ be the observed counts for a specific answer $a \in \mathcal{A}$ in the intervention and base conditions, respectively, such that $\sum_{a \in \mathcal{A}} C'_{S_i}(a) = N$. Additive smoothing with a parameter $\alpha$ is applied to derive smoothed probabilities, $p_{\alpha}$ and $q_{\alpha}$, from these counts:
$$
p_{\alpha}(a) = \frac{C'_{S_i}(a) + \alpha}{N + K\alpha} \quad \text{and} \quad q_{\alpha}(a) = \frac{C_{S_i}(a) + \alpha}{N + K\alpha}
$$
The smoothed KL divergence, $D_{\text{KL}}^{\alpha}$, is then computed using these non-zero probabilities:
$$
D_{\text{KL}}^{\alpha}[ p(A'_{S_i}) || p(A_{S_i}) ] = \sum_{a \in \mathcal{A}} p_{\alpha}(a) \log\left(\frac{p_{\alpha}(a)}{q_{\alpha}(a)}\right)
$$
This method replaces the use of a small $\epsilon$ floor. The smoothing parameters used are $\alpha = 1.0$ (Laplace smoothing) and $\alpha = 0.5$ (Jeffreys prior).

\begin{figure}
  \centering
  \includegraphics[width=1\linewidth]{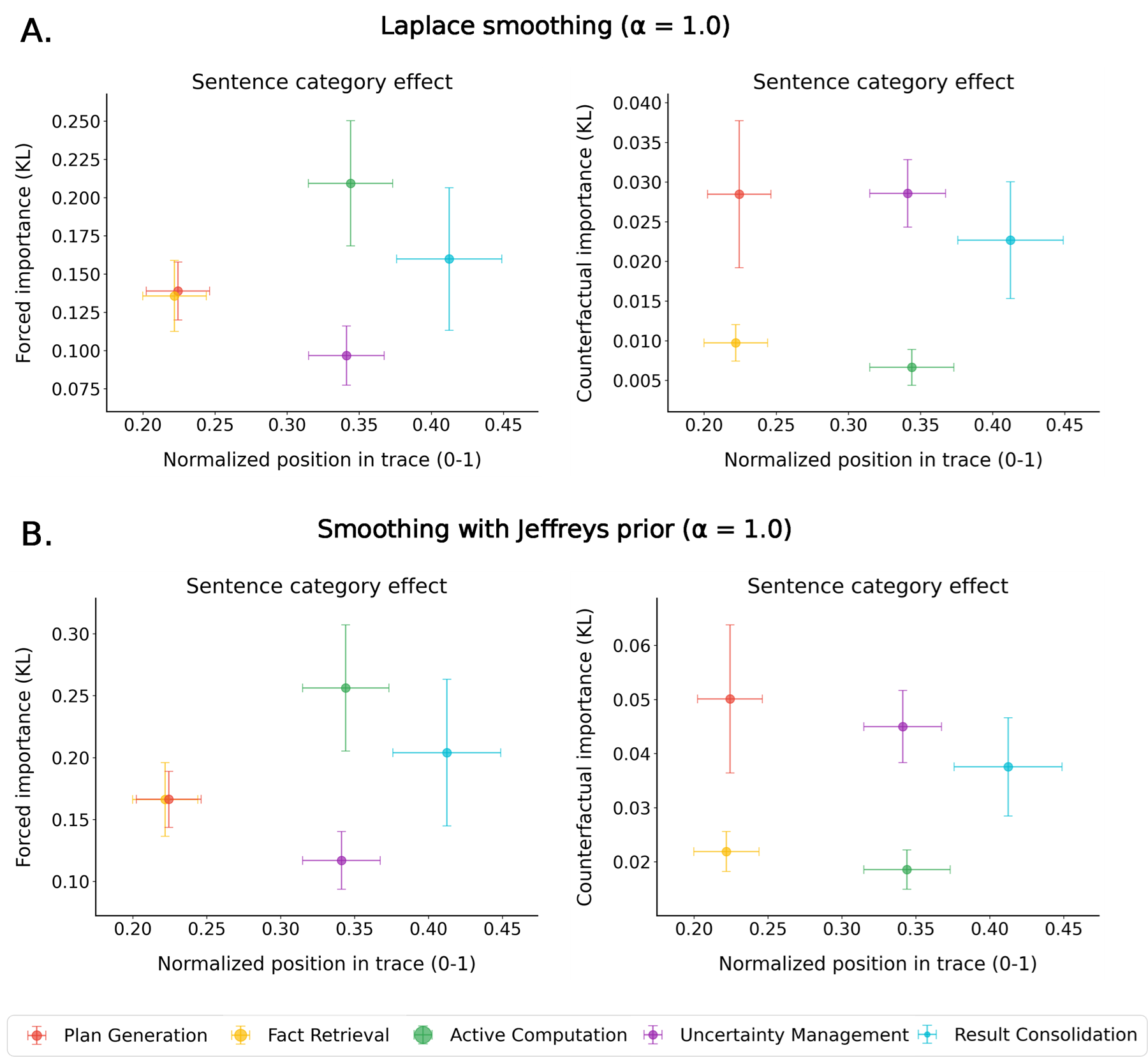}
  \caption{This is a variant of Figure~\ref{figure:sentence_category_importances}, now performed with smoothing. Smoothing was performed using either \textbf{(A)} $\alpha = 1.0$, Laplace smoothing, or \textbf{(B)} $\alpha = 0.5$, Jeffreys prior.}
  \label{figure:smoothing}
\end{figure}

With either level of smoothing, the same patterns linking importance and sentence category emerge as initially reported without smoothing (Figure~\ref{figure:sentence_category_importances}). Specifically, \textit{active computation} sentences yield higher forced answer importance than \textit{plan generation} and \textit{uncertainty management}, but the reverse is true when examining counterfactual importance based on the resampling method  (Figure~\ref{figure:smoothing}). 

\section{Generalizing to an alternative reasoning model}
\label{appendix-other-models}

\subsection{Measuring counterfactual influence}

To assess the generalizability of our counterfactual importance findings, we replicated our resampling methodology on R1-Distill-Llama-8B, applying the same experimental parameters (e.g., temperature = $0.6$ and top-p = $0.95$) used for R1-Distill-Qwen-14B. We collected 100 rollouts for 20 correct and 20 incorrect base solutions using the identical question set described in Section~\ref{secBlackBoxFramework}. 

\begin{figure}
  \centering
  \includegraphics[width=1\linewidth]{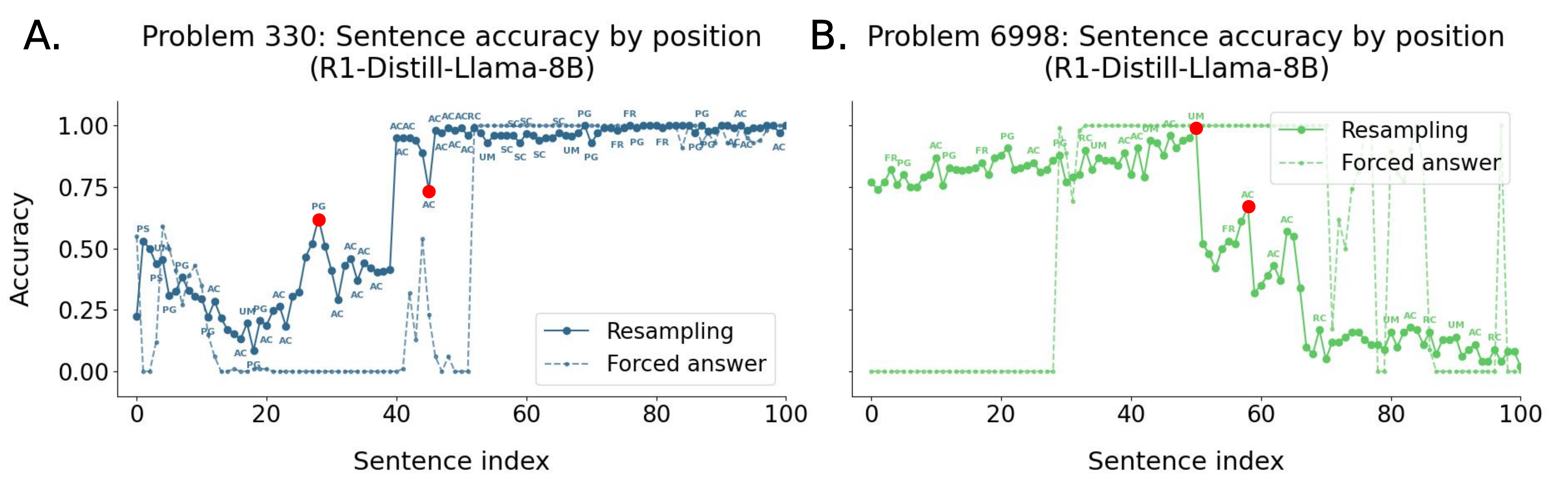}
  \caption{Accuracy over 100 rollouts at each sentence for (\textbf{A}) one correct and (\textbf{B}) one incorrect base solution for R1-Distill-Llama-8B. Red dots mark significant spikes or dips. Local minima and maxima sentences are annotated with category initials. Our analyses focus on the counterfactual KL-divergence between sentences, but resampling accuracy is visualized here as it is more intuitive.}
  \label{figure:resampling_examples_llama}
\end{figure}

The resampling accuracy trajectories for R1-Distill-Llama-8B (Figure~\ref{figure:resampling_examples_llama}) demonstrate patterns that are similar to those observed in R1-Distill-Qwen-14B (Figure~\ref{figure:resampling_examples}). Specifically, we observe similar characteristic accuracy fluctuations throughout the reasoning traces, with notable spikes and dips occurring at sentences corresponding to critical reasoning transitions.

\begin{figure}
  \centering
  \includegraphics[width=0.875\linewidth]{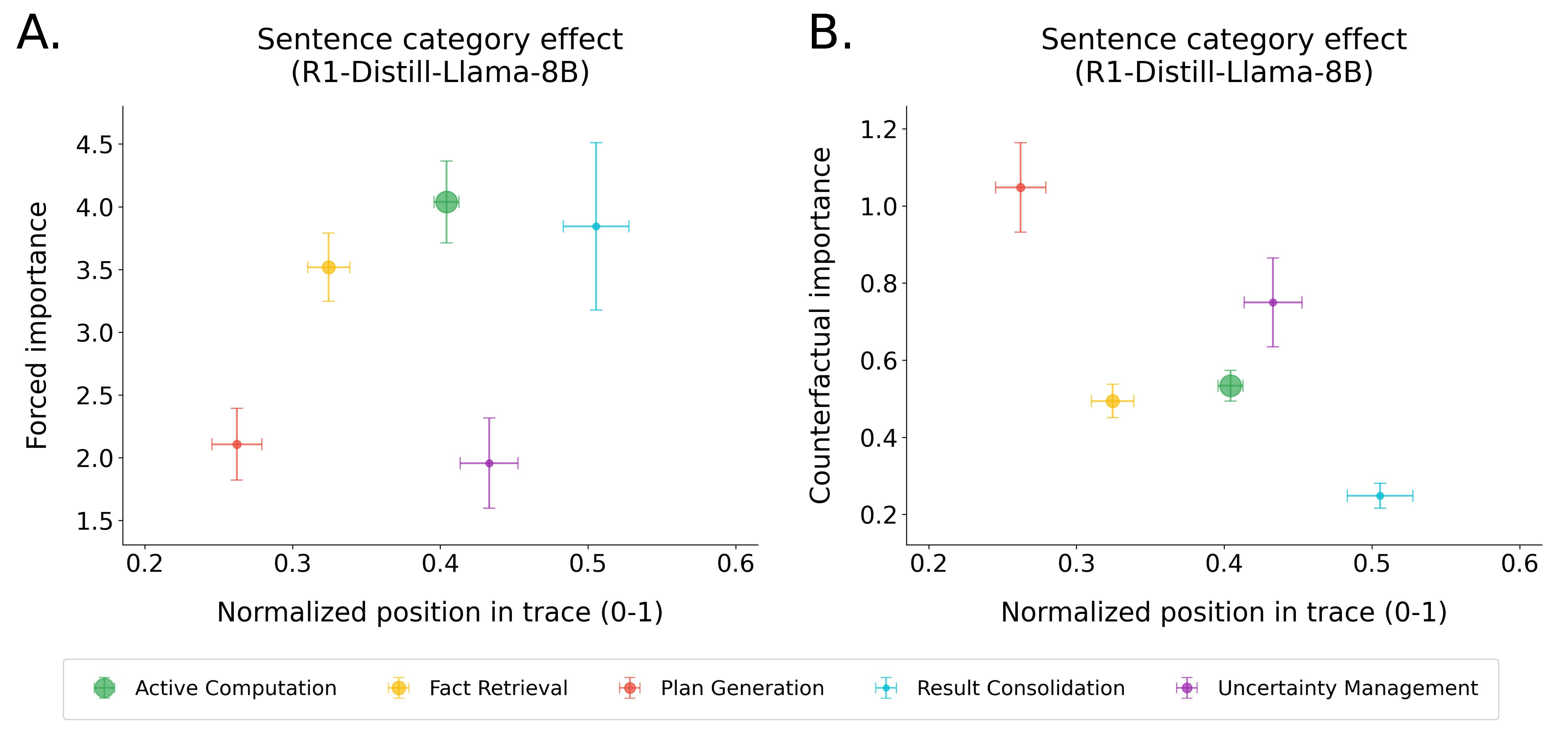}
  \caption{The mean of each sentence category for (\textbf{A}) forced-answer importance and (\textbf{B}) counterfactual importance for R1-Distill-Llama-8B, per the resampling method, plotted against the sentence category's mean position in the reasoning trace. Only the 5 most common sentence types are shown.}
  \label{figure:app_cat_eff_llama}
\end{figure}

Figure~\ref{figure:app_cat_eff_llama} shows that R1-Distill-Llama-8B exhibits similar sentence category effects whereby \textit{plan generation} and \textit{uncertainty management} sentences demonstrate higher counterfactual importance compared to \textit{active computation} and \textit{fact retrieval} sentences (see Figure ~\ref{figure:sentence_category_importances} for R1-Distill-Qwen-14B).

This cross-model validation supports our claim that reasoning traces are structured around high-level organizational sentences rather than low-level computational steps. The consistency of counterfactual importance patterns suggests that our sentence-level attribution framework captures fundamental properties of chain-of-thought reasoning that generalize beyond specific model implementations.

\subsection{Attention aggregation}

\begin{figure}[h]
  \centering
  \includegraphics[width=0.8\linewidth]{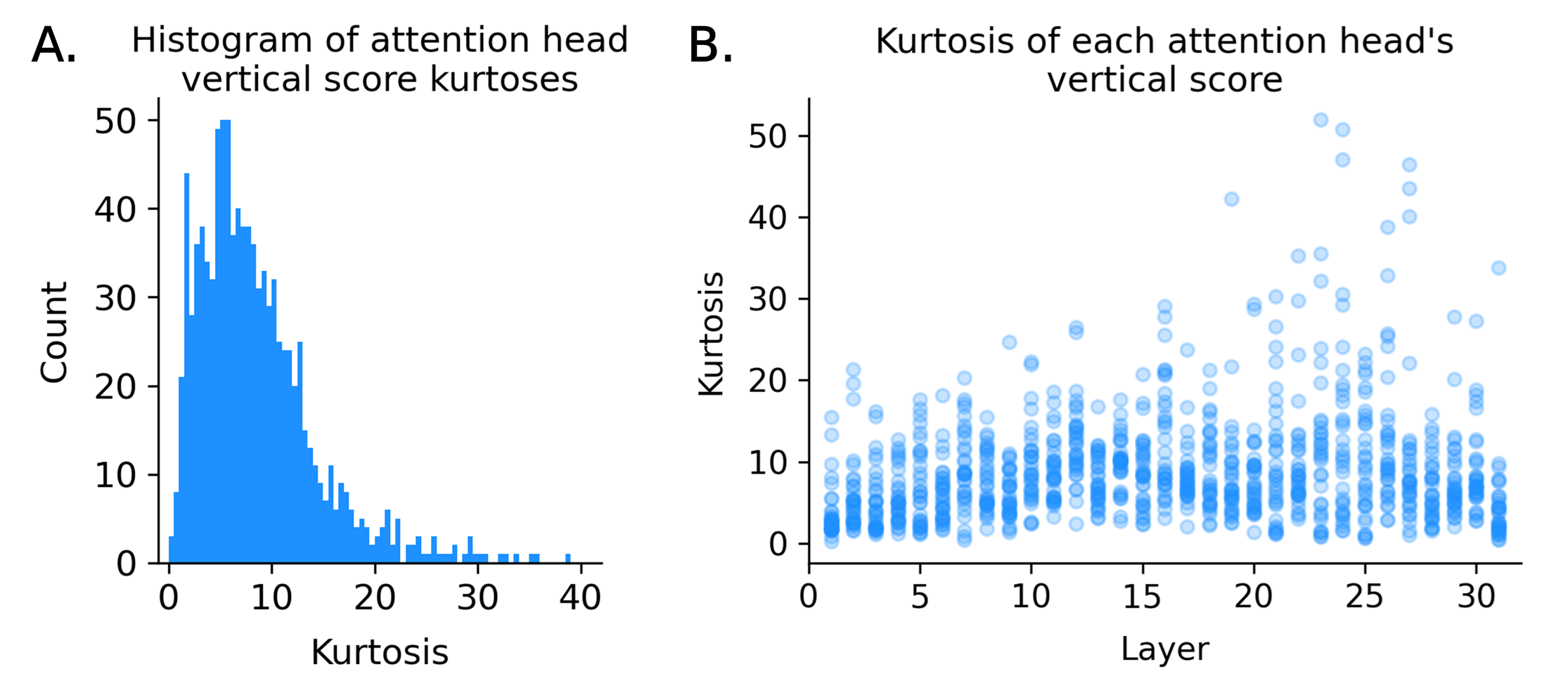}
  \caption{The plots here show the vertical-attention score patterns associated with the R1-Distill-Llama-8B data. \textbf{A.} This histogram shows the kurtosis values across all attention heads, median across all reasoning traces; parallels Figure~\ref{figure:rec_dist} based on the R1-Qwen-14B data. \textbf{B.} This scatterplot shows the kurtosis of each head's vertical-attention score, organized by layer. Figure~\ref{figure:rec_layer_kurt} is the R1-Distill-Qwen-14B version of this figure, which showed an upward trend into later layers that is not evident here.}
  \label{figure:rec_layer_kurt_llama}
\end{figure}

\begin{figure}[h]
  \centering
  \includegraphics[width=0.9\linewidth]{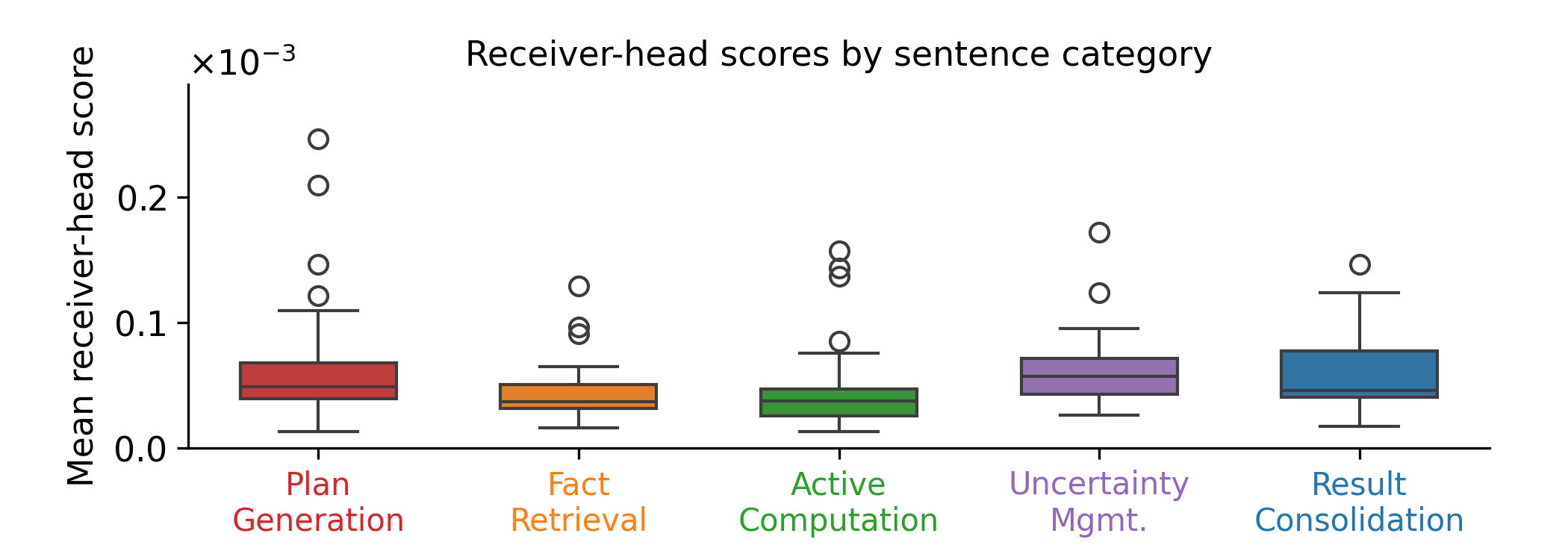}
  \caption{Based on the R1-Distill-Llama-8B data, the boxplot shows the average top-64 receiver-head score for each sentence type. The boxes correspond to the interquartile range across different reasoning traces. Figure~\ref{figure:rec_tax} is the R1-Distill-Qwen-14B version of this figure; note that for the R1-Distill-Qwen-14B figure, the top-16 heads were used. We found that for Llama 8B, examining the top-64 heads yielded more pronounced differences, although the sentence types with the highest scores remain the same.}
  \label{figure:rec_tax_llama}
\end{figure}

R1-Distill-Llama-8B displayed receiver-head patterns largely consistent with those of R1-Distill-Qwen-14B. The histogram of attention heads' vertical-attention scores displays a right tail, indicating that some attention heads tend to particularly focus attention on a subset of sentences (Figure~\ref{figure:rec_tax_llama}A). Interestingly, the R1-Distill-Qwen-14B receiver-heads tended to be more frequent in later layers (see below, Figure~\ref{figure:rec_layer_kurt}), which was not evident in R1-Distill-Llama-8B (Figure~\ref{figure:rec_layer_kurt_llama}). 

The R1-Distill-Qwen-14B and R1-Distill-Llama-8B receiver heads displayed consistent patterns related to sentence types, such that \textit{plan generation}, \textit{uncertainty management}, and \textit{self checking} sentences received heightened attention; although visually, the differences to \textit{fact retrieval} and \textit{active computation} may be less prominent, paired t-tests (paired with respect to a given response) showed that \textit{plan generation} and \textit{uncertainty management} always significantly surpassed \textit{fact retrieval} and \textit{active computation} (four paired t-tests: $p$s $\leq .01$).

No R1-Distill-Llama-8B results are provided for the attention suppression analysis, as that method was principally used for the case study, and no new case study was performed for R1-Distill-Llama-8B.

\section{Case study details}
\label{appendix-case-study}
\subsection{Case study transcript}
\label{appendix-case-transcript}

Below is a transcript of the model's response used for the case study (\Cref{secCaseStudy}). The first 74 sentences following \texttt{<think>} are shown. After this point, the model converged on 100\% resampling accuracy for all subsequent sentences. 

\begin{enumerate}
\setcounter{enumi}{-1}
\item (\texttt{problem\_setup}): Okay, so I have this problem where I need to find out how many bits the base-16 number 66666$_{16}$ has when it's converted to base-2.
\item (\texttt{uncertainty\_management}): Hmm, let's see.
\item (\texttt{fact\_retrieval}): I remember that each hexadecimal digit corresponds to exactly 4 binary digits, or bits.
\item (\texttt{plan\_generation}): So, maybe I can just figure out how many hexadecimal digits there are and multiply by 4 to get the number of bits.
\item (\texttt{uncertainty\_management}): Let me check that.
\item (\texttt{fact\_retrieval}): First, the number given is 66666 in base 16.
\item (\texttt{active\_computation}): Let me count the digits: 6, 6, 6, 6, 6.
\item (\texttt{result\_consolidation}): That's five digits in total.
\item (\texttt{active\_computation}): So, if each digit is 4 bits, then 5 digits would be 5×4=20 bits.
\item (\texttt{uncertainty\_management}): Wait, is that always the case?
\item (\texttt{fact\_retrieval}): I think so because each hex digit maps directly to 4 bits without any overlap or sharing.
\item (\texttt{result\_consolidation}): So, 66666$_{16}$ would be 20 bits in binary.
\item (\texttt{uncertainty\_management}): But wait, I should make sure that this is correct by actually converting the number or at least checking if there's any leading zero that might affect the bit count.
\item (\texttt{plan\_generation}): Alternatively, maybe I can calculate the value of 66666$_{16}$ in decimal and then find out how many bits that number would require.
\item (\texttt{plan\_generation}): Let's try that approach to cross-verify.
\item (\texttt{fact\_retrieval}): So, to convert 66666$_{16}$ to decimal, I can expand it as follows:
\item (\texttt{fact\_retrieval}): Each digit represents a power of 16.
\item (\texttt{fact\_retrieval}): Starting from the right, the digits are $16^{0}$,$16^{1}$,$16^{2}$,$16^{3}$,$16^{4}$.
\item (\texttt{fact\_retrieval}): So, the number is:
\item (\texttt{active\_computation}): 6×$16^{4}$+6×$16^{3}$+6×$16^{2}$+6×$16^{1}$+6×$16^{0}$
\item (\texttt{plan\_generation}): Let me compute each term:
\item (\texttt{fact\_retrieval}): First, $16^{4}$ is 16×16×16×16.
\item (\texttt{active\_computation}): $16^{2}$ is 256, so $16^{3}$ is 256×16=4096, and $16^{4}$ is 4096×16=65536.
\item (\texttt{active\_computation}): So, 6×65536=393216
\item (\texttt{active\_computation}): Next, $16^{3}$ is 4096, so 6×4096=24576
\item (\texttt{active\_computation}): Then, $16^{2}$ is 256, so 6×256=1536
\item (\texttt{active\_computation}): $16^{1}$ is 16, so 6×16=96
\item (\texttt{active\_computation}): And $16^{0}$ is 1, so 6×1=6
\item (\texttt{result\_consolidation}): Now, adding all these up:
\item (\texttt{active\_computation}): 393216+24576=417,792
\item (\texttt{active\_computation}): 417,792+1536=419,328
\item (\texttt{active\_computation}): 419,328+96=419,424
\item (\texttt{active\_computation}): 419,424+6=419,430
\item (\texttt{result\_consolidation}): So, 66666$_{16}$ is equal to 419,430 in decimal.
\item (\texttt{fact\_retrieval}): Now, to find out how many bits this number requires, I can use the formula for the number of bits needed to represent a number N in binary, which is $floor(log_{2}(N))+1$.
\item (\texttt{plan\_generation}): So, let's compute $log_{2}(419,430)$.
\item (\texttt{uncertainty\_management}): Hmm, I don't have a calculator here, but I can estimate it.
\item (\texttt{fact\_retrieval}): I know that $2^{18}$ =262,144
\item (\texttt{fact\_retrieval}): $2^{19}$ =524,288
\item (\texttt{result\_consolidation}): So, 419,430 is between $2^{18}$ and $2^{19}$.
\item (\texttt{result\_consolidation}): Therefore, $log_{2}(419,430)$ is between 18 and 19.
\item (\texttt{active\_computation}): The floor of that would be 18, so the number of bits is 18+1=19 bits.
\item (\texttt{uncertainty\_management}): Wait, but earlier I thought it would be 20 bits because it's 5 hex digits each being 4 bits.
\item (\texttt{result\_consolidation}): Now, according to this, it's 19 bits.
\item (\texttt{uncertainty\_management}): There's a discrepancy here.
\item (\texttt{uncertainty\_management}): I must have made a mistake somewhere.
\item (\texttt{self\_checking}): Wait, let me check my calculations again.
\item (\texttt{uncertainty\_management}): Maybe I messed up the decimal conversion.
\item (\texttt{fact\_retrieval}): Wait, 66666 in hex.
\item (\texttt{plan\_generation}): Let me convert it step by step again.
\item (\texttt{active\_computation}): 6×$16^{4}$: 6×65536=393216
\item (\texttt{active\_computation}): 6×$16^{3}$: 6×4096=24576
\item (\texttt{active\_computation}): 6×$16^{2}$: 6×256=1536
\item (\texttt{active\_computation}): 6×$16^{1}$: 6×16=96
\item (\texttt{active\_computation}): 6×$16^{0}$: 6×1=6
\item (\texttt{active\_computation}): Adding them up: 393216+24576=417,792
\item (\texttt{active\_computation}): 417,792+1536=419,328
\item (\texttt{active\_computation}): 419,328+96=419,424
\item (\texttt{active\_computation}): 419,424+6=419,430.
\item (\texttt{result\_consolidation}): That seems correct.
\item (\texttt{result\_consolidation}): So, 419,430 in decimal.
\item (\texttt{fact\_retrieval}): Now, let's check $2^{18}$ is 262,144, $2^{19}$ is 524,288.
\item (\texttt{result\_consolidation}): So, 419,430 is less than 524,288, so $log_{2}(419,430)$ is less than 19.
\item (\texttt{active\_computation}): So, $floor(log_{2}(419,430))$ is 18, so number of bits is 19.
\item (\texttt{uncertainty\_management}): But wait, when I thought about each hex digit being 4 bits, 5 hex digits would be 20 bits.
\item (\texttt{uncertainty\_management}): So, why is there a difference?
\item (\texttt{plan\_generation}): Ah, perhaps because leading zeros are not counted.
\item (\texttt{fact\_retrieval}): So, when converting from hex to binary, each digit is 4 bits, but if the leading hex digits are such that when converted to binary, they don't add any leading zeros beyond the 4 bits per digit.
\item (\texttt{uncertainty\_management}): Wait, but 66666 in hex is 5 digits, so 5×4=20 bits.
\item (\texttt{result\_consolidation}): But according to the decimal conversion, it's 19 bits.
\item (\texttt{result\_consolidation}): That suggests that when converted to binary, it's 19 bits, not 20.
\item (\texttt{uncertainty\_management}): So, which is correct?
\item (\texttt{plan\_generation}): Wait, perhaps when converting $66666_{16}$ to binary, it's exactly 20 bits, but the leading bits are zeros, so when counting the number of bits, we don't count leading zeros.
\item (\texttt{fact\_retrieval}): Wait, but in the problem, it's just asking how many bits it has when written in base 2.
\end{enumerate}

\subsection{Case study: Sentence 13 alternatives}
\label{appendix-sentence-13-alternatives}

Sentence 13 (\textit{``Alternatively, maybe I can calculate the value of $66666_{16}$ in decimal and then find out how many bits that number would require.''}) was found to have the highest counterfactual importance among any sentence in the response (see Figure ~\ref{figure:resampling_examples}A). Five alternative possible sentences resampled at the sentence 13 position, marked by whether they eventually led to a correct or incorrect answer, are as follows:

\begin{enumerate}
    \item ($\times$) Let me think. The first digit is 6, which in binary is 0110.
    \item ($\checkmark$) Let me try converting the number to decimal first\dots
    \item ($\times$) Let me try converting the first few digits to binary to see how it goes.
    \item ($\times$) Let me think about the conversion process.
    \item ($\checkmark$) Let me try converting the number step by step.
\end{enumerate}

\subsection{Receiver head and sentence-sentence case study findings}
\label{appendix-case-depth}

\begin{figure}
  \centering
  \includegraphics[width=1.0\linewidth]{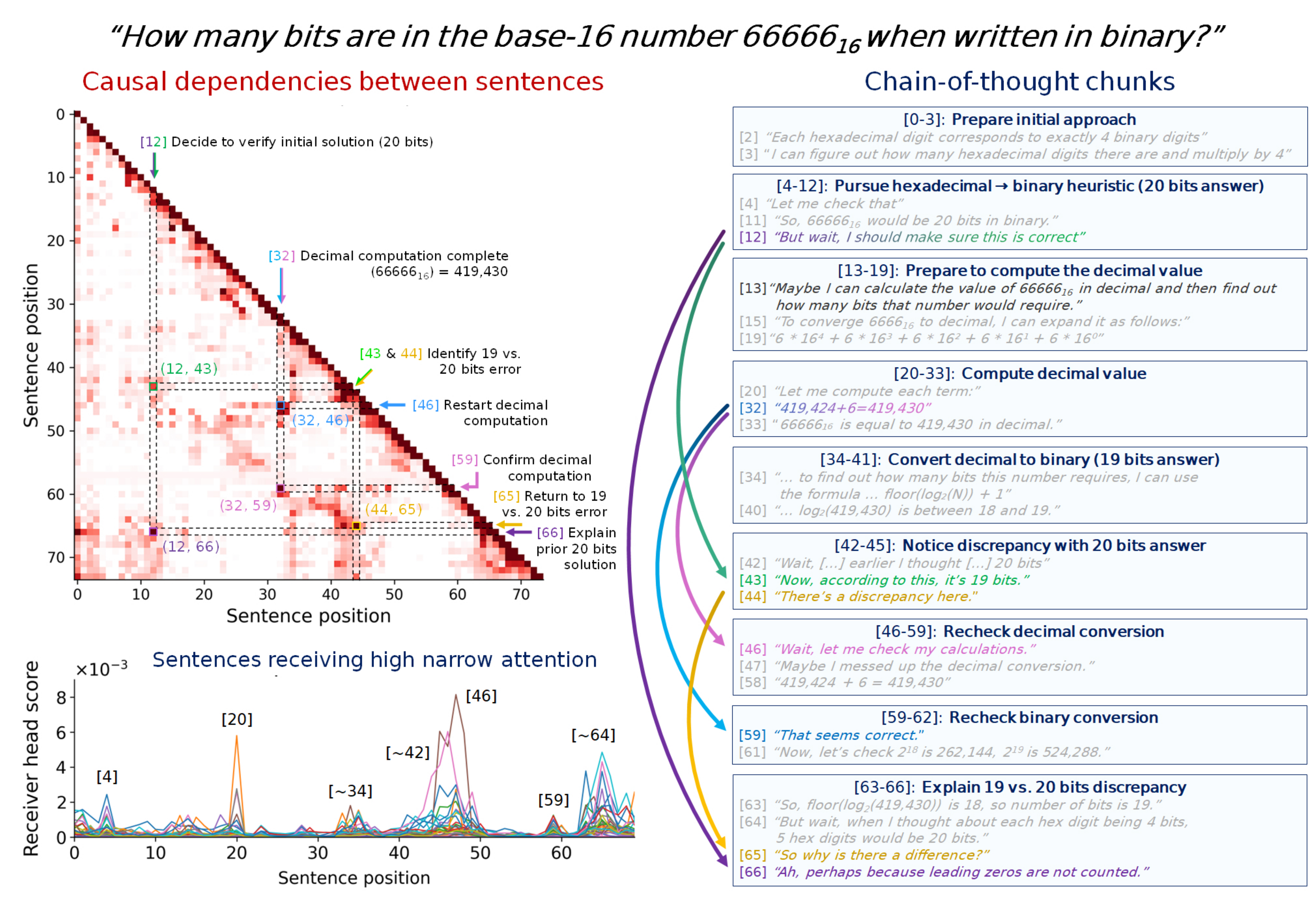}
  \caption{Case study: problem \#4682 (correct). Red matrix shows the effect of suppressing one sentence (x-axis) on a future sentence (y-axis). Darker colors indicate higher values. Bottom-left line plot shows the average attention toward each sentence by all subsequent sentences via the top-32 receiver heads (32 attention heads with the highest kurtosis score). Flowchart summarizes the model’s CoT with chunks defined around key sentences receiving high attention via receiver heads. Sentence 13 is emphasized as it has high counterfactual importance per the resampling method (see Figure~\ref{figure:resampling_examples}A).}
  \label{figure:case}
\end{figure}

The presented techniques cover different aspects of attribution within a reasoning trace. Building on the case-study conclusions from our resampling approach (\cref{case-resampling}), we study the model's CoT here by focusing on receiver heads and sentence-sentence links (Figure~\ref{figure:case}) (see above, \Cref{appendix-case-study}, for the full transcript). 

\subsubsection{Receiver heads}
 
The trajectory toward the final correct answer can be understood as a series of computational chunks (see flowchart in Figure~\ref{figure:case}). First, the model prepares a formula for converting $66666_{16}$ to decimal (sentences 13-19). Next, the model computes the answer to that formula, finding that $66666_{16}$ is 419,430 in decimal (sentences 20-33). The model subsequently converts that number to binary by putting forth another formula and solving it, $floor(log_{2}(419,430)) + 1 = 19$, to derive that the answer is ``19 bits'' (sentences 34-41). The model then notes a discrepancy with the earlier 20-bit solution (sentences 42-45). The model hence initiates new computations that verify that it computed the decimal value of $66666_{16}$ correctly (sentences 46-58) and that it computed the binary conversion accurately (sentences 59-62). Equipped with this increased certainty about 19-bit answer, the model discovers why its initial 20-bit idea was incorrect: \textit{``because leading zeros are not counted''} (Sentence 66). This overall narrative is based on our analysis of attention patterns (\cref{secAttentionAggregation}): Receiver attention heads pinpoint sentences initiating computations or stating key conclusions, thereby segmenting the reasoning trace into seemingly meaningful chunks (Figure~\ref{figure:case}).

\subsubsection{Attention suppression}

Along with being organized into computational chunks, the reasoning displays a scaffold related to sentence-sentence dependencies (Figure~\ref{figure:case}). One piece of this structure is a self-correction pattern involving an incorrect proposal, a detected discrepancy, and a final resolution. Specifically, the model initially proposes an incorrect answer of “20 bits”, which it decides to recheck (sentence 12). This leads to a discrepancy with the “19 bits” answer computed via decimal conversion (sentences 43 \& 44). After rechecking its arithmetic supporting the “19 bit” answer, the model returns to the discrepancy (sentence 65) and then produces an explanation for why the “20 bits” answer is incorrect (sentence 66). This can be seen as a tentative CoT circuit, where two conclusions conflict to produce a discrepancy, which in turn encourages the model to resolve the discrepancy. Within this wide-spanning scaffold, there exist further dependencies, corresponding to verifying an earlier computation. Specifically, the model finishes computing the decimal value of $66666_{16}$ as 419,430 (sentence 32), later decides to verify that decimal conversion (sentence 46), and finally confirms that the original value is correct (sentence 59). This can be seen as further indication of CoT circuitry.

We identified these linkages based on the attention-suppression matrix (\cref{sec:attention-suppression}), which contains local maxima at these linkages (12 → 43, 43 → 65, 12 → 66; 32 → 46, 32 → 59). Notice that many of the sentences pinpointed by the attention-suppression technique overlap with the sentences receiving high attention from receiver heads. Adding to the receiver-head conclusions, the attention suppression technique shows how information flows between these key sentences that structure the reasoning trace.

\subsection{Sentence position effects on receiver-head scores}
\label{appendix-position-rec}

A sentence’s position within the reasoning trace will tend to influence its measured receiver score.

As a reasoning trace progresses, the number of possible broadcasted sentences will necessarily increase. For instance, by sentence 20, there might be only two broadcasted sentences (each receiving 50\% of attention from sentences 21-29), whereas by sentence 100, there could be ten broadcasted sentences (each receiving 10\% of attention from sentences 101-109). As the sum of an attention weight row will sum to 1 (at the token level), later sentences will distribute their attention across a larger number of past sentences. This dilution of attention creates downward pressure on the receiver-head scores of later sentences. This is the case even though a receiver head score extends through all subsequent low-competition or high-competition periods. For example, broadcasting sentence 20 will face limited competition from receiving sentence 21-29 attention and high competition for sentences 101-109, whereas broadcasting sentence 100 will exclusively face high competition, pushing its score downward as broadcasting-sentence position increases.

There also exists a proximity effect on receiver-head scores that operates in the opposite direction of the above effect. Although broadcasted sentences are attended by all subsequent sentences to some degree, this will be more so the case for more recently subsequent sentences (e.g., receiving more attention from a sentence 10 sentences downstream than one 20 sentences downstream). For sentences late in the reasoning trace, the average distance to future sentences will be shorter. For example, if a reasoning trace contains 120 sentences, then sentence 100 will be at most 19 sentences apart from any given future sentence, whereas sentence 20 will be at most 99 sentences apart. To a degree, the analyses in the report account for proximity effects by ignoring the 4 sentences immediately proximal to a given sentence when calculating vertical-attention scores. However, this will not fully address proximity effects.

We see no reason why the downward pressure of sentence position on receiver-head scores (attention dilution) will be equal in magnitude to the upward pressure of sentence position (proximity effects).

For the preparation of the present report, we conducted exploratory analyses evaluating whether the above confounding factors invalidate any presented finding, and we did not find evidence that this is the case. Thus, rather than pursuing some technique to account for the above pressures (e.g., linearly weighing attention weight matrices based on their position), we opted to only account for these factors in a minimal fashion by ignoring the attention among sentences just 4 sentences apart.

\section{Sentence taxonomy}
\label{appendix-taxonomy}

Building on top of the framework presented by \citep{venhoff2025understanding}, we developed a taxonomy consisting of eight distinct sentence categories that capture reasoning functions in mathematical problem-solving. Each category represents a specific cognitive operation. The functions and examples for each category are given in Table~\ref{table:sentence_taxonomy}. Notably, the \textit{uncertainty management} category includes backtracking sentences.

\begin{table}[H]
  \caption{Sentence taxonomy with reasoning functions in problem-solving}
  \label{table:sentence_taxonomy}
  \centering
  \begin{tabular}{p{3.8cm}p{5cm}p{4cm}}
    \toprule
    \textbf{Category} & \textbf{Function} & \textbf{Examples} \\
    \midrule
    \textbf{Problem Setup} & Parsing or rephrasing the problem (e.g., initial reading) & \textit{I need to find the area of a circle with radius 5 cm.} \\
    \midrule
    \textbf{Plan Generation} & Stating or deciding on a plan of action, meta-reasoning & \textit{I'll solve this by applying the area formula.} \\
    \midrule
    \textbf{Fact Retrieval} & Recalling facts, formulas, problem details without computation & \textit{The formula for the area of a circle is $A = \pi r^2$.} \\
    \midrule
    \textbf{Active Computation} & Algebra, calculations, or other manipulations toward the answer & \textit{Substituting $r = 5$: $A = \pi \times 5^2 = 25\pi$.} \\
    \midrule
    \textbf{Uncertainty Management} & Expressing confusion, re-evaluating, including backtracking & \textit{Wait, I made a mistake earlier. Let me reconsider...} \\
    \midrule
    \textbf{Result Consolidation} & Aggregating intermediate results, summarizing, or preparing & \textit{So the area is $25\pi$ square cm which is approximately...} \\
    \midrule
    \textbf{Self Checking} & Verifying previous steps, checking calculations, and re-confirmations & \textit{Let me verify: $\pi r^2 = \pi \times 5^2 = 25\pi$. Correct.} \\
    \midrule
    \textbf{Final Answer Emission} & Explicitly stating the final answer & \textit{Therefore, the answer is...} \\
    \bottomrule
  \end{tabular}
\end{table}

\begin{figure}[H]
  \centering
  \includegraphics[width=1\linewidth]{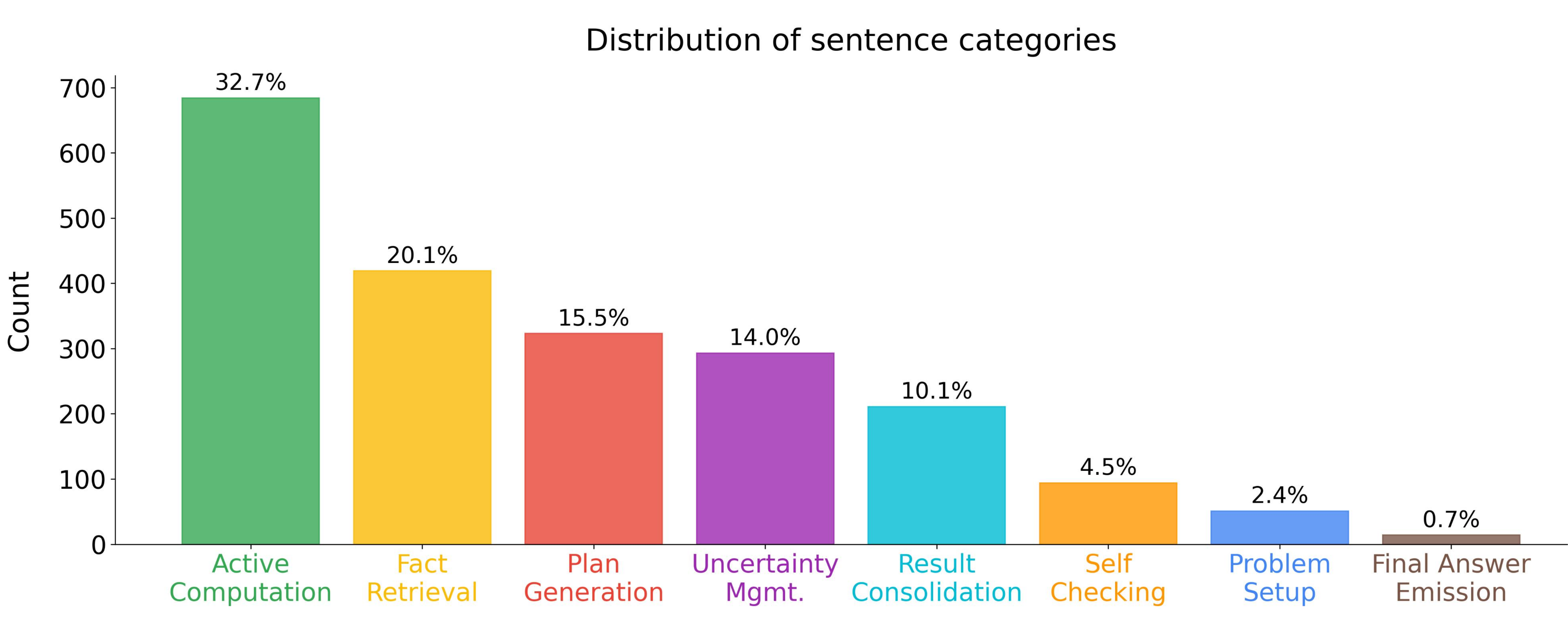}
  \caption{Counts and frequencies of taxonomic sentence categories in our dataset.}
  \label{figure:app_cat_freqs}
\end{figure}

The distribution of categories across our dataset as shown in Figure~\ref{figure:app_cat_freqs} reveals that \textit{active computation} constitutes the largest proportion (32.7\%), followed by \textit{fact retrieval} (20.1\%), \textit{plan generation} (15.5\%), and \textit{uncertainty management} (14.0\%). The sequential structure of reasoning is reflected in the rarity and positioning of \textit{problem setup} (2.4\%), which typically occurs at the beginning, and \textit{final answer emission} (0.7\%), which predominantly appears toward the end of the reasoning process.

\newpage
\section{Prompt information}
\label{appendix-for-prompt}
We used the following prompt with OpenAI GPT-4o (April-May, 2025) to annotate each sentence:\\\\\\
{
\tt You are an expert in interpreting how LLMs solve math problems using multi-step reasoning. Your task is to analyze a chain-of-thought reasoning trace, broken into discrete text sentences, and label each sentence with:

1. **function\_tags**: One or more labels that describe what this sentence is *doing* functionally in the reasoning process.\\
2. **depends\_on**: A list of earlier sentence indices that this sentence directly depends on, e.g., uses information, results, or logic introduced in earlier sentences.\\\\
This annotation will be used to build a dependency graph and perform causal analysis, so please be precise and conservative: only mark a sentence as dependent on another if its reasoning clearly uses a previous sentence's result or idea.\\\\
Function Tags:\\\\
1. problem\_setup: Parsing or rephrasing the problem (initial reading or comprehension).\\
2. plan\_generation: Stating or deciding on a plan of action (often meta-reasoning).\\
3. fact\_retrieval: Recalling facts, formulas, problem details (without immediate computation).\\
4. active\_computation: Performing algebra, calculations, manipulations toward the answer.\\
5. result\_consolidation: Aggregating intermediate results, summarizing, or preparing final answer.\\
6. uncertainty\_management: Expressing confusion, re-evaluating, proposing alternative plans (includes backtracking).\\
7. final\_answer\_emission: Explicit statement of the final boxed answer or earlier sentences that contain the final answer. \\
8. self\_checking: Verifying previous steps, checking calculations, and re-confirmations.\\
9. unknown: Use only if the sentence does not fit any of the above tags or is purely stylistic or semantic.\\\\
Dependencies:\\\\
For each sentence, include a list of earlier sentence indices that the reasoning in this sentence *uses*. For example:\\
- If sentence 9 performs a computation based on a plan in sentence 4 and a recalled rule in sentence 5, then depends\_on: [4, 5]\\
- If sentence 24 plugs in a final answer to verify correctness from sentence 23, then depends\_on: [23]\\
- If there's no clear dependency use an empty list: []\\
- If sentence 13 performs a computation based on information in sentence 11, which in turn uses information from sentence 7, then depends\_on: [11, 7]\\\\
Important Notes:\\
- Make sure to include all dependencies for each sentence.\\
- Include both long-range and short-range dependencies.\\
- Do NOT forget about long-range dependencies.\\
- Try to be as comprehensive as possible.\\
- Make sure there is a path from earlier sentences to the final answer.\\
Output Format:\\\\
Return a dictionary with one entry per sentence, where each entry has:\\
- the sentence index (as the key, converted to a string),\\
- a dictionary with:\\
    \text{     } - "function\_tags": list of tag strings\\
    \text{     } - "depends\_on": list of sentence indices, converted to strings\\\\
Here is the expected format:\\
\{\\
    \text{     }"1": \{\\
    \text{     }        "function\_tags": ["problem\_setup"],\\
    \text{     }        "depends\_on": [""]\\
    \text{     }\},\\
    \text{     }"4": \{\\
    \text{     }        "function\_tags": ["plan\_generation"],\\
    \text{     }        "depends\_on": ["3"]\\
    \text{     }\},\\
    \text{     }"5": \{\\
    \text{     }        "function\_tags": ["fact\_retrieval"],\\
    \text{     }        "depends\_on": []\\
    \text{     }\},\\
    \text{     }"9": \{\\
    \text{     }        "function\_tags": ["active\_computation"],\\
    \text{     }        "depends\_on": ["4", "5"]\\
    \text{     }\},\\
    \text{     }"24": \{\\
    \text{     }        "function\_tags": ["uncertainty\_management"],\\
    \text{     }        "depends\_on": ["23"]\\
    \text{     }\},\\
    \text{     }"32": \{\\
    \text{     }        "function\_tags": ["final\_answer\_emission"],\\
    \text{     }        "depends\_on": ["9, "30", "32"]\\
    \text{     }\},\\
\}\\\\
Here is the math problem:\\
<PROBLEM>\\\\
Here is the full chain-of-thought, broken into sentences:\\
<SENTENCES>\\\\
Now label each sentence with function tags and dependencies.
}

\newpage

\section{Sentence category probing}
\label{appendix-probing}

\begin{figure}[H]
  \centering
  \includegraphics[width=1\linewidth]{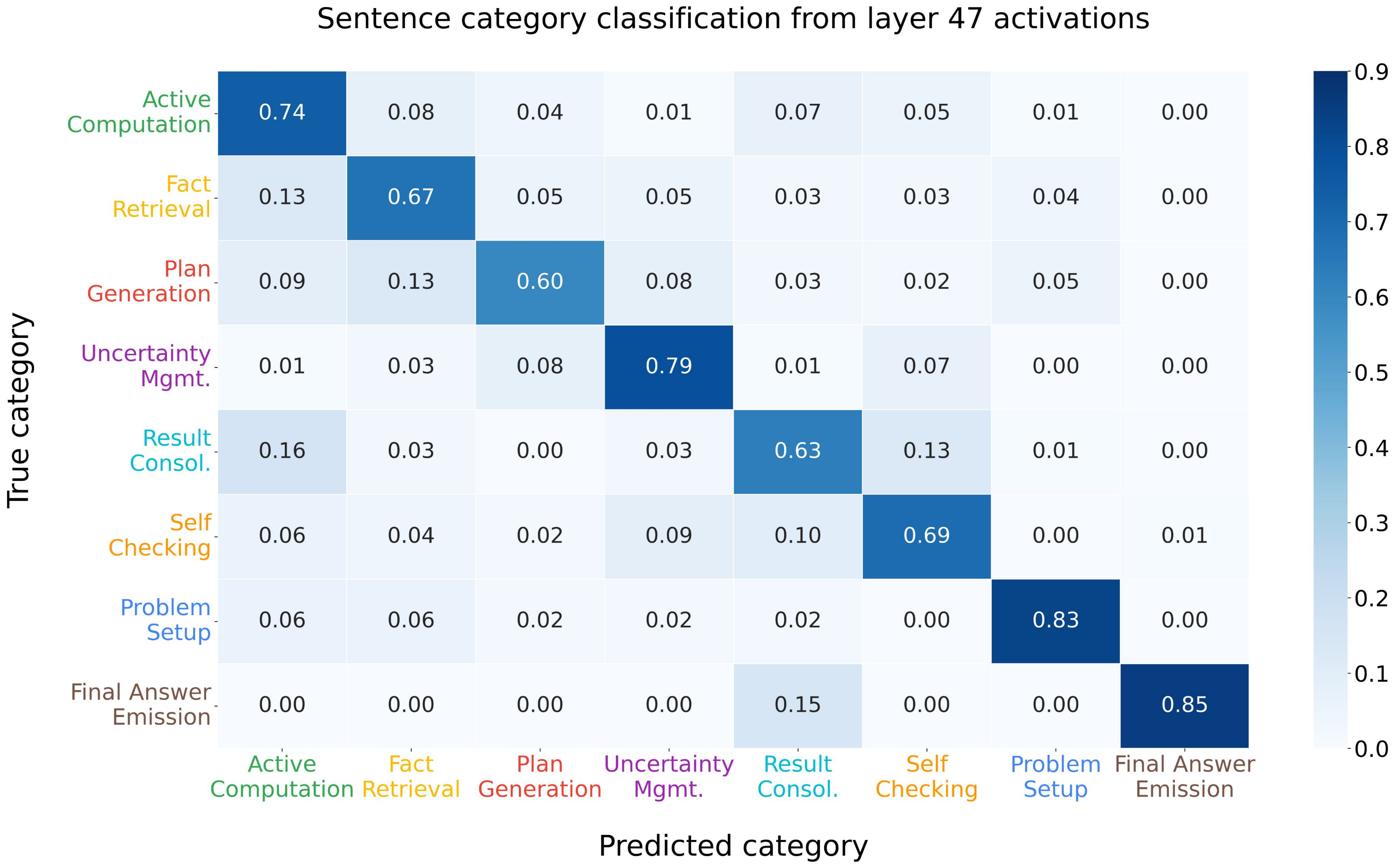}
  \caption{Confusion matrix showing the sentence category classification performance of a logistic regression probe trained on activations from layer 47 of the R1-Distill-Qwen-14B model. Values represent the proportion of examples from each true category (rows) classified as each predicted category (columns). Diagonal elements indicate correct classifications.}
  \label{figure:app_cat_prob}
\end{figure}

We trained a linear classifier to identify sentence categories based on activations. We employed a multinomial logistic regression with L2 regularization ($C=1.0$) on the residual stream activity from layer $47$ (last layer) of R1-Distill-Qwen-14B. For evaluating accuracy, we implemented a group-5-fold cross-validation that ensured examples from the same problem response remained in either the training or testing set to prevent data leakage. We averaged the residual stream activity across tokens to create sentence-level representations, whose dimensions were then standardized. To address class imbalance in the training data, we employed balanced class weights. The model demonstrated strong discriminative power across all reasoning categories, achieving a macro-F1 score of $0.71$. The confusion matrix presented in Figure~\ref{figure:app_cat_prob} reveals high classification accuracies for categories such as \textit{active computation} ($0.74$), \textit{uncertainty management} ($0.79$), and \textit{problem setup} ($0.83$), while showing some confusion between functionally related categories.

\section{Embeddings model}
\label{appEmbeddings}

We used \texttt{all-MiniLM-L6-v2} with a maximum sequence length of $256$ tokens and a hidden dimension of $384$ as our sentence embeddings model from the \texttt{sentence-transformers} \citep{reimers-2019-sentence-bert} library. We picked a cosine similarity threshold of $0.8$, which is the median similarity value between all sentence removed (i.e., original sentence) and sentence resampled pairs in our dataset.

\section{Additional resampling results}
\label{appendix-resampling}

\begin{figure}[H]
  \centering
  \includegraphics[width=0.885\linewidth]{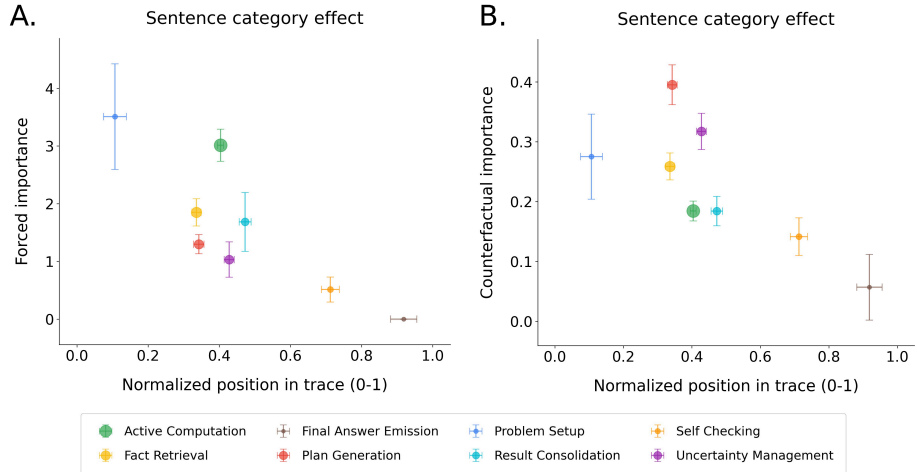}
  \caption{The mean of each sentence category for (\textbf{A}) forced-answer importance and (\textbf{B}) counterfactual importance for R1-Distill-Qwen-14B, per the resampling method, plotted against the sentence category's mean position in the reasoning trace. All sentence types are shown.}
  \label{figure:app_res_res}
\end{figure}

Figure~\ref{figure:app_res_res} presents mean counterfactual importances across all eight taxonomic categories for R1-Distill-Qwen-14B, extending the main text results (Figure~\ref{figure:sentence_category_importances}) which showed only the five most frequent sentence types. The expanded view includes three additional categories with lower frequencies. \textit{Problem setup} sentences occur predominantly at trace beginnings (mean normalized position $\approx 0.1$) with moderate-high counterfactual importance. \textit{Self checking} sentences tend to occur in the second-half of the traces and show lower counterfactual importance. \textit{Final answer emission} sentences appear late in traces (mean normalized position $\approx 0.9$) and show the lowest counterfactual importance. The patterns observed in the five-category analysis remain consistent when examining the full taxonomy.

\section{Counterfactual versus resampling importance} 
\label{appendix-resample-methods}


The resampling importance metric introduced in Section~\ref{subsecCounterfactualImportance} treats all resampled sentences as equally informative, but different sentence types may exhibit varying degrees of \textbf{overdetermination} during resampling. Overdetermination occurs when resampled sentences $T_i$ are frequently similar to the original sentence $S_i$ (i.e., $T_i \approx S_i$), indicating that the reasoning context strongly constrains what can be expressed at that position. We present empirical evidence that counterfactual importance is a more nuanced measure by accounting for semantic divergence in resampled content.

\begin{figure}[H]
  \centering
  \includegraphics[width=1\linewidth]{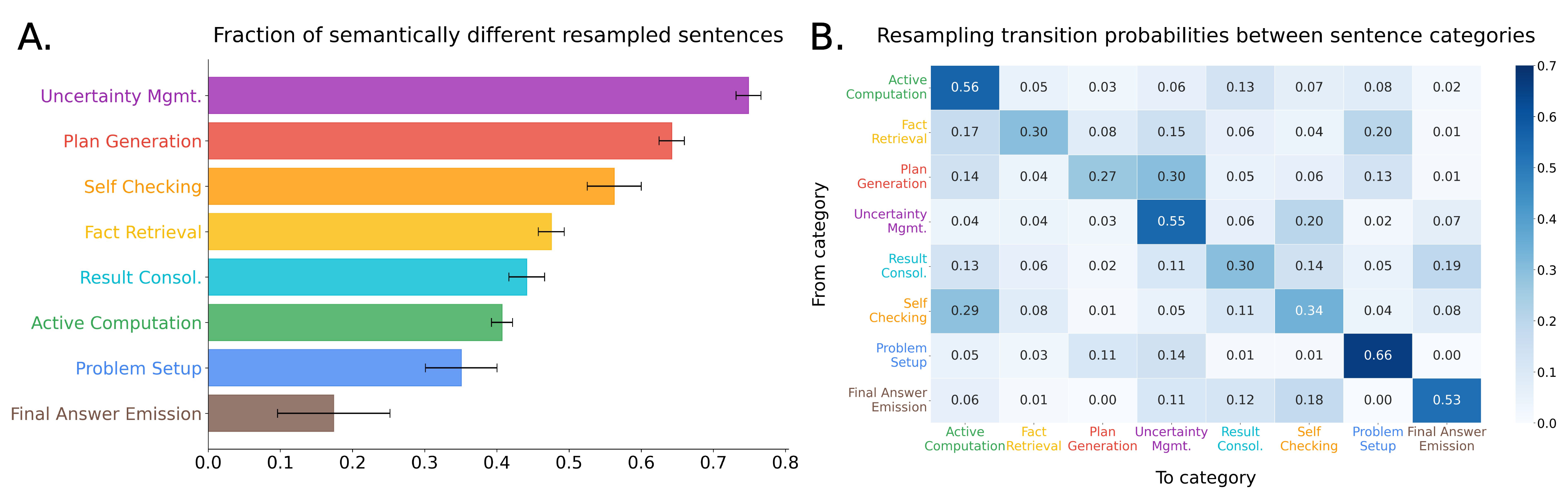}
  \caption{(\textbf{A}) Fraction of semantically different resampled sentences by category, showing that \textit{uncertainty management} and \textit{plan generation} sentences produce more divergent alternatives when resampled. (\textbf{B}) Transition probabilities between original and resampled sentence categories.}
  \label{figure:app_od}
\end{figure}

Some sentences are more overdetermined than others. Figure~\ref{figure:app_od}A shows that \textit{uncertainty management} and \textit{plan generation} sentences produce semantically different alternatives in a large proportion of resamples, while \textit{active computation} and \textit{problem setup} sentences show lower divergence rates.

The transition matrix in Figure \ref{figure:app_od}B shows how sentence categories change under resampling. For instance, \textit{uncertainty management} and \textit{active computation} sentences are usually replaced by sentences of the same category, whereas \textit{plan generation} and \textit{fact retrieval} sentences are more often resampled into a variety of other categories.

\begin{figure}[H]
  \centering
  \includegraphics[width=1\linewidth]{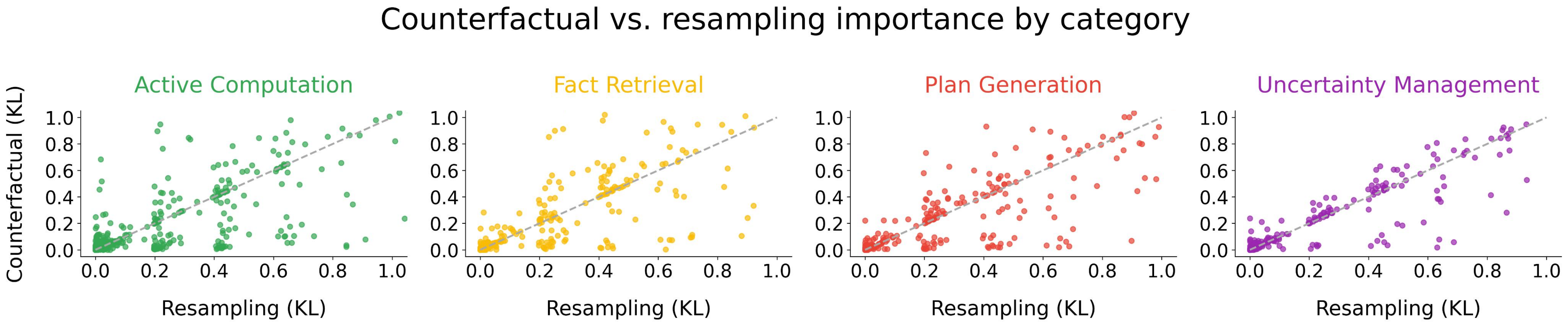}
  \caption{Comparison between counterfactual and resampling importance metrics across sentence categories. Each point represents a single sentence and the dashed gray line is the $y = x$ line.}
  \label{figure:app_coun_vs_res}
\end{figure}

These resampling behaviors create systematic differences between our counterfactual and resampling importance metrics. Figure~\ref{figure:app_coun_vs_res} demonstrates that the relationship between the two metrics varies substantially across sentences and sentence categories. The counterfactual importance metric aims to address overdetermination by explicitly filtering for semantically different resamples, providing a more targeted measure of causal influence. In contrast, the resampling metric potentially overestimates the importance of sentences that consistently produce similar content when resampled.

However, the counterfactual importance metric can yield high-variance estimates when the number of semantically divergent resampled sentences is low (e.g., $<10$), as the conditional probability estimates become less reliable with limited data. Alongside the limitations discussed in Section~\ref{secDiscussion}, this represents another constraint of our approach that future work should investigate further.

\section{Additional receiver head information}
\label{appendix-receiver-heads}

\begin{figure}
  \centering
  \includegraphics[width=0.55\linewidth]{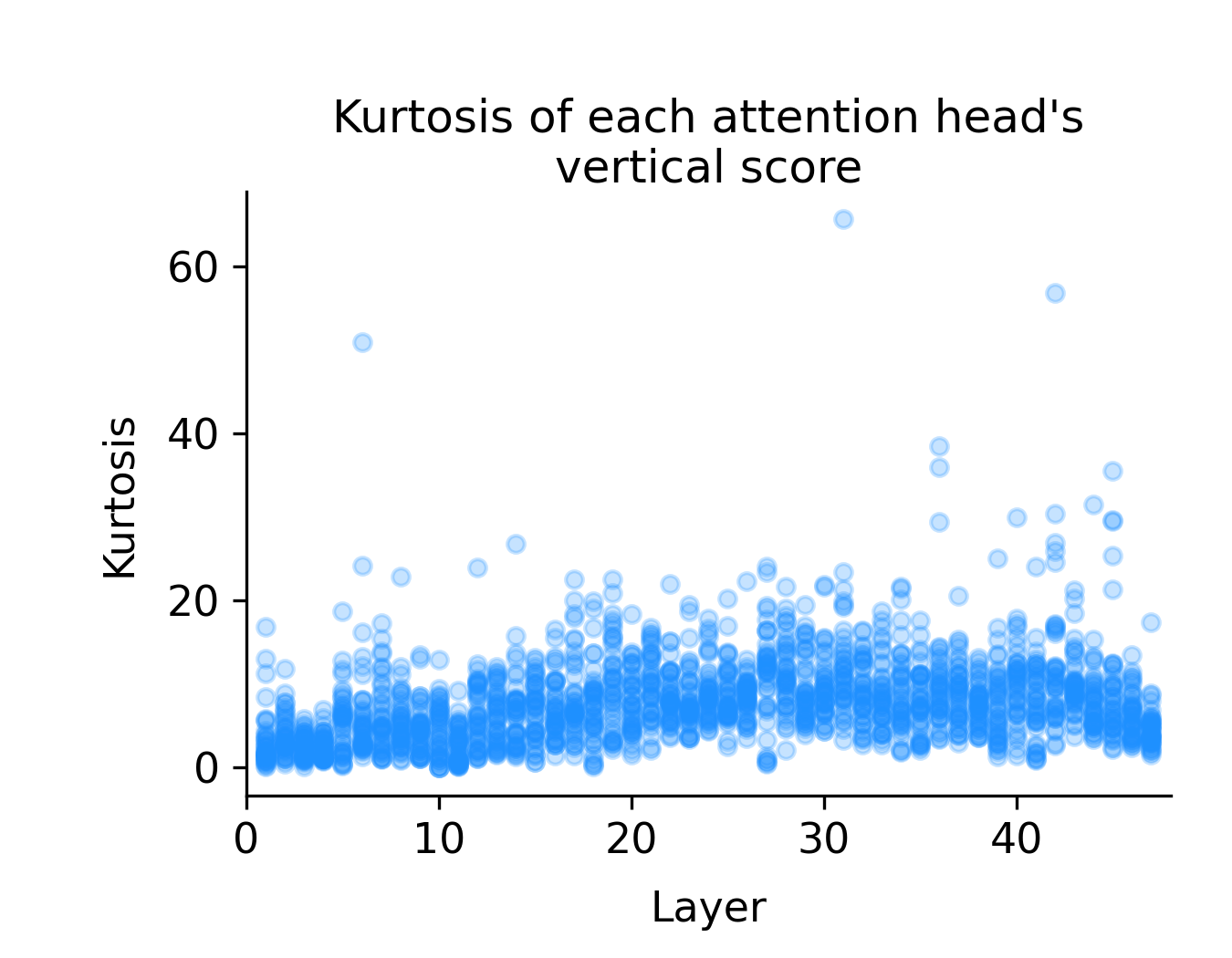}
  \caption{This scatterplot shows the kurtosis of each head's vertical-attention score, organized by layer. There is an upward trend across layers and a strong uptick among some late-layer heads.}
  \label{figure:rec_layer_kurt}
\end{figure}

\begin{figure}
  \centering
  \includegraphics[width=0.85\linewidth]{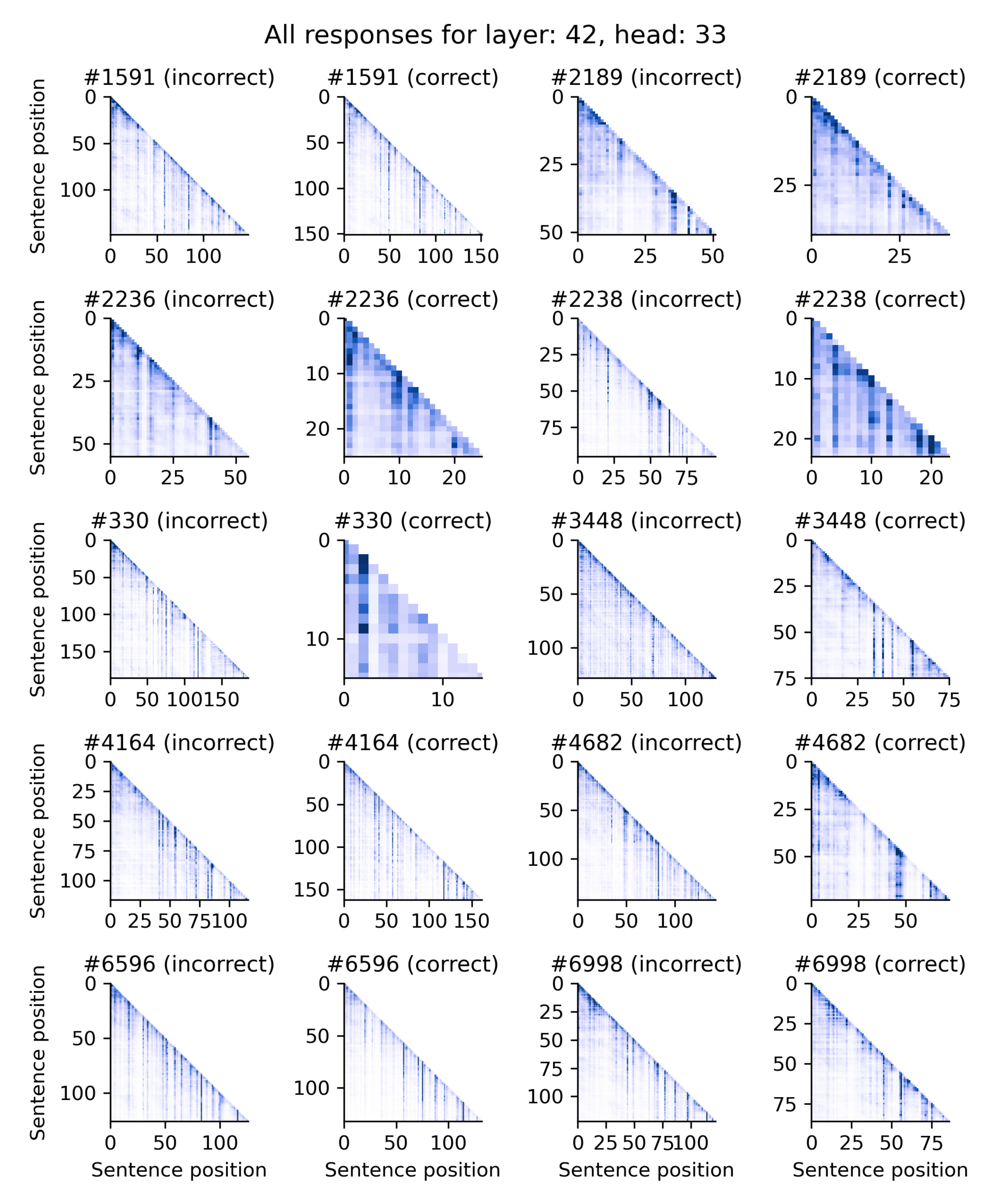}
  \caption{The attention weight matrices for the receiver head with the highest kurtosis score are shown here for twenty of the forty responses (selected arbitrarily based on the first twenty processed). The coloring was defined such that the darkest navy corresponds to values surpassing 99.5th percentile value of each matrix. White is zero.}
  \label{figure:rec_many_pn}
\end{figure}

\begin{figure}
  \centering
  \includegraphics[width=0.85\linewidth]{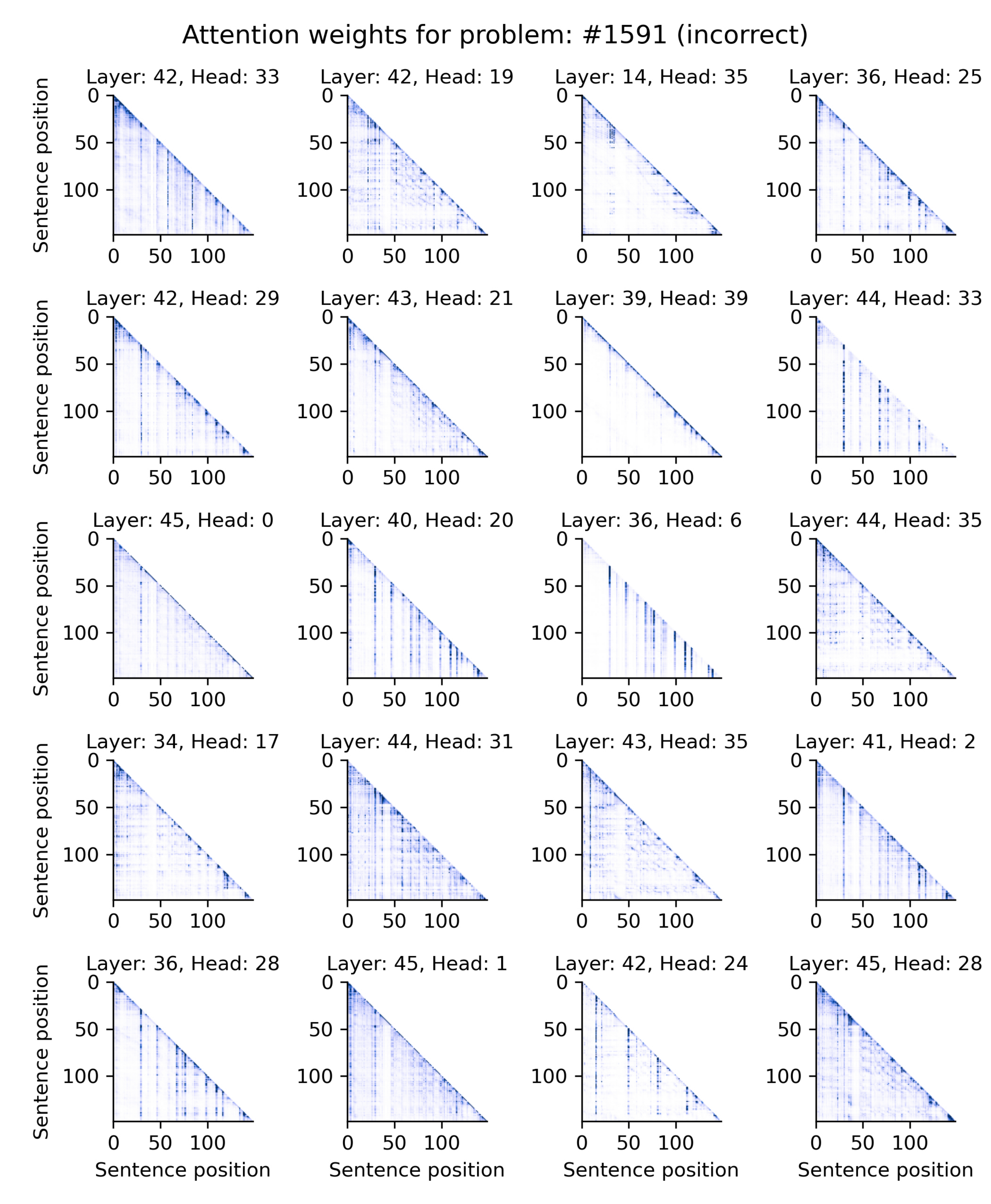}
  \caption{The attention weight matrices for response \#1591 (incorrect) are shown here for the 20 attention heads yielding the highest kurtosis score across all responses. No effort was taken to ``cherry-pick'' responses showing prominent receiver head patterns; we are showing \#1591 (incorrect) because it corresponded to the alphabetically earliest problem number among the ten problems analyzed (correct/incorrect chosen randomly). The coloring was defined such that the darkest navy corresponds to values surpassing 99.5th percentile value of each matrix. White is zero.}
  \label{figure:rec_many_head}
\end{figure}

Receiver heads -- heads receiving high kurtosis scores -- are more common in late layers (Figure~\ref{figure:rec_layer_kurt}). Examples of receiver heads are shown in Figure~\ref{figure:rec_many_pn}, showing how the highest kurtosis head consistently narrows attention on particular sentences, and Figure~\ref{figure:rec_many_head}, showing how there exist many heads that narrow attention on particular sentences. 

\section{Reasoning versus base model differences in receiver heads}
\label{rec-head-difs}
\begin{figure}
  \centering
  \includegraphics[width=0.85\linewidth]{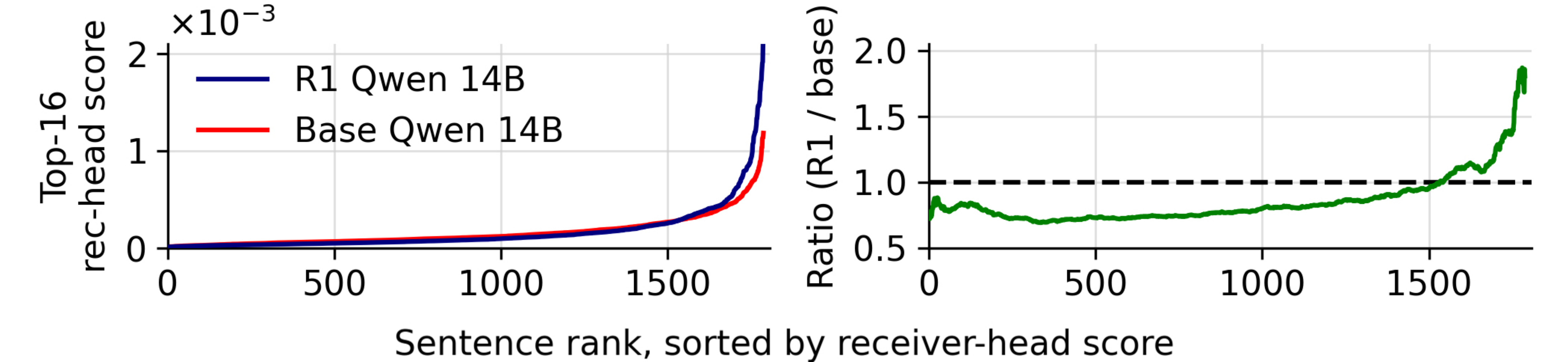}
  \caption{The navy and red lines on the left show the receiver-head scores assigned to sentences, averaged across the 16 heads with the highest kurtoses. The green lines on the right represent the ratio of the navy and blue lines for a given sentence rank. Sentences with high receiver head scores receive more attention in the reasoning model compared to the base model.}
  \label{figure:rec_dif}
\end{figure}

Attentional narrowing toward particular sentences may be a feature specifically of reasoning models. We submitted the reasoning traces to a base model version of Qwen-14B and identified receiver heads. For both models, we sorted all sentences by their mean receiver-head score using the 16 attention heads with the highest kurtoses. The highest percentile sentences received greater attention by the reasoning model - e.g., the highest-percentile sentences receive 1.8x more attention via top-16 heads in the reasoning model compared to the base model (Figure~\ref{figure:rec_dif}). Additionally, lower percentile sentences receive less attention through the top-16 heads. This conclusion is somewhat tenuous, as no base-model difference is seen when this result is tested using R1-Distill-Llama-8B. Nonetheless, based on the Qwen-14B data, it appears the model has learned to narrow its attention toward particular sentences.

\section{Effects of ablating receiver heads}
\label{appendix-ablation}

To test the causal hypothesis that the receiver heads identified in \Cref{secAttentionAggregation} are functionally important for reasoning, we performed an experiment ablating receiver heads and evaluating how this impact's model accuracy. This intervention is designed to measure the direct impact of removing these heads on task performance and to evaluate the possibility that they may be more important than typical heads.

\subsection{Methodology}

We continue to use problems from the MATH dataset. We selected 32 problems where the non-ablated model achieves 10-90\% accuracy on average. For each problem, we ran R1-Distil-Qwen-14B sixteen times, while allowing the model to output up to $2^{16}$ (16,384) tokens. Responses that did not produce an answer by that point were marked as incorrect.

We compared the effect of ablating 128 attention heads (approx. 7\% of all heads), 256 heads (approx. 13\%), or 512 heads (approx. 27\%). The ablation strategies were:

\begin{compactenum}
    \item \textbf{Receiver head ablation:} We ablated the top-N heads with the highest average kurtosis scores.
    \item \textbf{Random non-receiver (control) ablation:} For each layer where $k$ receiver heads were ablated, we ablated $k$ heads chosen randomly from the set of heads not selected from that same layer. This ensures a matched comparison with no overlap. 
\end{compactenum}

Note that receiver heads are more common in late layers (see above, Figure~\ref{figure:rec_layer_kurt}). By ensuring that both conditions included an equal number of heads from each layer (rather than selecting 128, 256, or 512 heads randomly across all layers), this ensures that differences cannot be explained simply by differences in the layers selected.

In the 512-head ablation condition, a majority of attention heads in some late layers were marked as receiver heads. For these layers, the non-receiver control condition was modified to ablate the corresponding number of heads with the lowest kurtosis scores to ensure a valid comparison set. For instance, if 60\% of layer 43 heads are in the top-512, then the control condition included the 60\% with the lowest kurtosis score, meaning that there is 20\% overlap for that layer.

\subsection{Results and Discussion}

Our experiments show that a large number of heads must be ablated to induce a significant drop in performance compared to the baseline level of accuracy (baseline = 64.1\%, 95\% CI: [56.0\%, 72.1\%]). Regardless of whether receiver heads or non-receiver heads are targeted, ablating 128 heads produces differences in accuracy that insignificantly differ from baseline accuracy, and ablating 256 heads still produces only a small drop in accuracy (Table \ref{table:ablation_results}).\footnote{We are not aware of prior studies on attention head ablation for models generating long chain-of-thought reasoning, making it difficult to establish what is a typical number of heads to ablate. Potentially, a large number is necessary because the long reasoning traces (sometimes exceeding 10,000 tokens) provide extensive opportunities for error correction and compensatory computation.}



\begin{table}[H]
  \caption{Answer accuracy on MATH problems for different self-attention-head ablation conditions. The brackets show the 95\% confidence interval for each accuracy estimate.}
  \label{table:ablation_results}
  \centering
  \begin{tabular}{cccc}
    \toprule
    \textbf{Heads Ablated} & \textbf{Receiver heads} & \textbf{Random heads} \\
    \midrule
    256 &  48.8\% [39.3\%, 58.3\%] & 52.7\% [43.0\%, 62.5\%]\\
    512 & 27.7\% [17.2\%, 38.2\%] & 37.3\% [27.5\%, 47.1\%] \\
    \bottomrule
  \end{tabular}
\end{table}

The importance of receiver heads emerges when a large number of heads are ablated. When ablating 512 heads (over a quarter of the model's 1920 heads), targeting receiver heads caused performance to fall to 28\% accuracy. Removing the same number of control heads resulted in a less severe drop to 37\% accuracy. There is a significant difference between these percentages ($t[31] = 2.55, p = .02$), suggesting receiver heads are more critical for reasoning than other heads.

As mentioned, this analysis treats responses as incorrect if they do not produce a final answer by 16,384 tokens. If the analysis is changed to instead simply omit those responses entirely from the analysis, there remains a significant difference in accuracy when ablating top-512 receiver heads (29\% accuracy) versus random non-receiver heads (39\% accuracy) ($t[31] = 2.66, p = .02)$. Hence, regardless of whether non-completed responses are marked as incorrect or ignored, ablating receiver heads is found to exert a larger impact on model accuracy than ablating random non-receiver heads.

\section{KL causal graph pseudocode}
\label{appendix-graph-pseudocode}

This pseudocode outlines the procedure for computing a sentence-to-sentence causal graph for a given chain-of-thought (CoT). The algorithm works by systematically masking each source sentence and measuring the resulting change in the model's predictions (logits) for all subsequent target sentences. The sentence--sentence impact is quantified as the average log-KL divergence across a target sentence's tokens, which is then normalized against the average impact from all prior sentences. This last normalization step effectively accounts for differences in target sentences' average entropy, which may vary widely and can hamper studying differences between target sentences.

Masking can be performed either by suppressing attention toward the source sentence or omitting the sentence entirely; the former preserves positional embedding information, while the latter may be computationally cheaper and easier to implement (e.g., with serverless providers). If masking is done by omitting sentence $i$ from the CoT, rather than suppressing attention toward sentence $i$, this will impact sentence $j$'s token positions across the CoT and masked CoT, which should be accounted for.

\begin{algorithm}[H]
\caption{GetCausalMatrix(CoT, Model)}
\begin{algorithmic}[1]
\State Initialize $\textsc{Causal\_Matrix}\in\mathbb{R}^{M\times M}\gets 0$ \Comment{$M$ = number of sentences in CoT}
\State $\textsc{logits\_base} \gets \textsc{Forward\_Pass}(\text{CoT}, \text{Model})$ \Comment{shape: (tokens, vocabulary)}
\For{$i = 0$ \textbf{to} $M-1$} \Comment{source sentence}
    \State $\text{CoT}_{\text{masked}} \gets \textsc{Mask\_Source}(\text{CoT}, i)$
    \State $\textsc{logits\_masked} \gets \textsc{Forward\_Pass}(\text{CoT}_{\text{masked}}, \text{Model})$
    \For{$j = i+1$ \textbf{to} $M-1$} \Comment{target sentence}
        \State $\textsc{tokens\_j} \gets \textsc{Sentence\_Tokens}(\text{CoT}_{\text{masked}}, j)$
        \State $\textsc{total\_KL} \gets 0$
        \For{\textbf{each} $k \in \textsc{tokens\_j}$}
            \State $\textsc{KL} \gets \textsc{KLDivergence}(\textsc{logits\_base}[k], \textsc{logits\_masked}[k])$
            \State $\textsc{total\_KL} \gets \textsc{total\_KL} + \log(\textsc{KL})$
        \EndFor
        \State $\textsc{Causal\_Matrix}[i, j] \gets \textsc{total\_KL} / |\textsc{tokens\_j}|$
    \EndFor
\EndFor
\State \Comment{Normalize each target column by the mean over prior sources}
\For{$j = 0$ \textbf{to} $M-1$}
    \State $\mu \gets \textsc{mean}(\textsc{Causal\_Matrix}[0\!:\!j,\, j])$
    \State $\textsc{Causal\_Matrix}[0\!:\!j,\, j] \gets \textsc{Causal\_Matrix}[0\!:\!j,\, j] - \mu$
\EndFor
\State \Return $\textsc{Causal\_Matrix}$
\end{algorithmic}
\end{algorithm}

\begin{algorithm}[H]
\caption{KLDivergence$(\textsc{logits\_p}, \textsc{logits\_q})$}
\begin{algorithmic}[1]
\State $\log p \gets \textsc{logits\_p} - \textsc{log\_sum\_exp}(\textsc{logits\_p})$ \Comment{log-softmax}
\State $\log q \gets \textsc{logits\_q} - \textsc{log\_sum\_exp}(\textsc{logits\_q})$
\State $p \gets \exp(\log p)$
\State $\textsc{kl} \gets 0$
\For{\textbf{each} vocabulary index $v$}
    \State $\textsc{kl} \gets \textsc{kl} + p[v]\cdot\big(\log p[v] - \log q[v]\big)$
\EndFor
\State \Return $\textsc{kl}$
\end{algorithmic}
\end{algorithm}

\section{Sentence-to-sentence counterfactual importance}
\label{appendix-sentence-sentence-counterfactual}

\label{Methodological details}

We extend our counterfactual resampling framework (\cref{subsecSentenceTaxonomy}) to quantify each sentence's influence on each future sentence. Further below, we describe how this measure's values for sentence-sentence links correlate with the values generated via our \cref{sec:attention-suppression} method, masking sentences and measuring the impact on later sentences' logits. 

We estimate the counterfactual importance of sentence $S_i$ on a future sentence $S^\text{Fut.}$ formally with: 

\begin{equation}
\text{importance}(S_i \to S^\text{Fut.}) = \mathbb{P}(S^\text{Fut.} \in_{\approx} \{ S_i, \dots , S_M \}) - \mathbb{P}(S^\text{Fut.} \in_{\approx} \{ T_i, \dots , T_N \} | T_i \not\approx S_i)
\label{eqnSentenceToSentence}
\end{equation}

Intuitively, on the right-hand side of \Cref{eqnSentenceToSentence}, the first term is the probability that a future sentence $S^{\text{Fut.}}$ will semantically occur given that $S_i$ was present in the trace, and the second term is the corresponding probability when $S_i$ is resampled with a non-equivalent sentence. A positive score indicates that sentence $S_i$ increases the likelihood of producing $S^{\text{Fut.}}$ (i.e.,  $S_i$ upregulates $S^{\text{Fut.}}$), while a negative score suggests that it suppresses or inhibits it.

In this context $S^{\text{Fut.}}$ semantically occurs if, when we extract the sentences and identify the best candidate match for $S^{\text{Fut.}}$ using cosine similarity between sentence embeddings, it has greater than $0.8$ cosine similarity (i.e., the median value in our dataset) to that sentence. Pseudocode for estimating sentence-to-sentence importance and empirical values of this metric can be found in \Cref{appendix-resample-methods}.

\begin{figure}[H]
  \centering
  \includegraphics[width=0.625\linewidth]{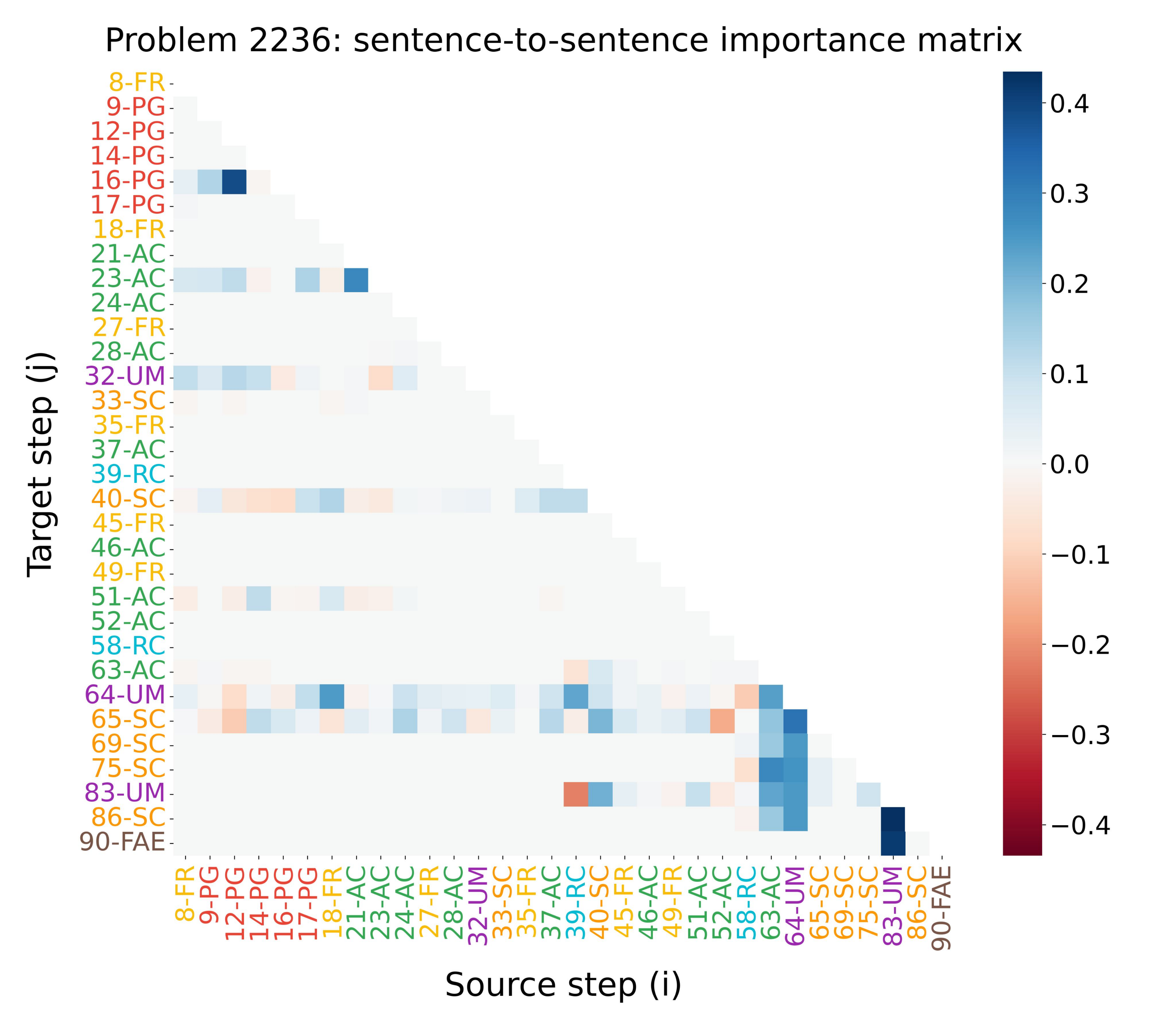}
  \caption{Sentence-to-sentence importance matrix for the 32 most important sentences in problem \#2236 (incorrect), selected based on total outgoing and incoming importance. Each cell $(i,j)$ shows the causal importance of sentence $i$ on sentence $j$, calculated as the difference in the probability sentence $j$ semantically occurs ($>0.8$ cosine similarity) when sentence $i$ is present versus resampled.}
  \label{figure:app_sen_sen}
\end{figure}

Beyond measuring individual sentence importance, our framework quantifies causal dependencies between specific sentence pairs within reasoning traces. Figure \ref{figure:app_sen_sen} displays the sentence-to-sentence importance matrix for problem \#2236 (incorrect) (\textit{``Each page number of a 488-page book is printed one time in the book. The first page is page 1 and the last page is page 488. When printing all of the page numbers, how many more 4's are printed than 8's?''}), showing how individual sentences influence downstream reasoning steps. Below we list a few illustrative cases.

\begin{itemize}
\item \textbf{12-PG $\rightarrow$ 16-PG.}
The planning in sentence 12 (\textit{``1. Count the number of 4's in the units place across all page numbers''}) raises the probability that the model produces sentence 16 (\textit{``Starting with the 4's.''}) by $0.39$. A plan statement triggers a subordinate planning step.

\item \textbf{8-FR, 9-PG, 12-PG, 14-PG $\rightarrow$ 32-UM.}
The uncertainty management in sentence 32 (\textit{``However, I need to check if 440-449 is fully included.''}) receives sizeable positive influence from several earlier sentences: 8-FR ($+0.11$), 9-PG ($+0.06$), 12-PG ($+0.12$), 14-PG ($+0.10$). This forms the dense horizontal band at row index 32.

\item \textbf{39-RC $\not \rightarrow$ 83-UM.}
The result consolidation in sentence 32 (\textit{``Now, summing up all the 4's: - Units: 48 - Tens: 50 - Hundreds: 89. Total 4's = 48 + 50 + 89 = 187.''}) decreases the likelihood (i.e., inhibits) of 83-UM (\textit{``Wait, but just to be thorough, let me check the hundreds place for 4’s again.''}) by $0.22$.

\item \textbf{52-AC $\not \rightarrow$ 65-SC.}
The computation in sentence 52 (\textit{``The first four blocks 80-89, 180-189, 280-289, 380-389 each contribute 10 eights in the tens place.''}) decreases the likelihood of 65-SC (\textit{``Let me go through each step again to make sure I didn't make a mistake.''}) by $0.16$.

\item \textbf{63-AC $\rightarrow$ 64-UM, 65-SC, 69-SC, 75-SC, 83-UM, 86-SC.}
The computation in sentence 63 (\textit{``So, the difference is 187 – 98 = 89.''}) propagates forward, increasing the likelihood of 64-UM ($+0.24$), 65-SC ($+0.17$), 69-SC ($+0.16$), 75-SC ($+0.28$), 83-UM ($0.23$), and 86-SC ($0.16$). This forms the dense vertical band originating from column index 63.

\item \textbf{64-UM $\rightarrow$ 65-SC, 69-SC, 75-SC, 83-UM, 86-SC.}
The uncertainty management in sentence 64 (\textit{``Wait, that seems quite a large difference.''}) further amplifies the same downstream block: 65-SC ($+0.32$), 69-SC ($+0.25$), 75-SC ($+0.26$), 83-UM ($0.25$), and 86-SC ($0.25$).

\item \textbf{83-UM $\rightarrow$ 86-SC, 90-FAE.}
Even very late checks matter. Sentence 83 (\textit{``Wait, but just to be thorough, let me check the hundreds place for 4’s again.''}) increases the chance of 86-SC (\textit{``Correct. And for the tens place...''}) by $0.43$ and of the final answer in 90-FAE by $0.41$.
\end{itemize}

We provide the following pseudocode for estimating sentence-to-sentence importance:

\begin{verbatim}
Input: Sentence index i, target sentence index j (where j > i), threshold t = 0.8
Output: Importance score importance(i -> j)

1. Get rollouts R_keep where sentence i was kept (resampling from i+1)
2. Get rollouts R_remove where sentence i was removed (resampling from i)

3. For each rollout r in R_keep:
   a. Extract all sentences S_r from rollout r
   b. Find best matching sentence to target sentence j:
      - Compute sentence embeddings
      - Calculate cosine similarity between each s in S_r and target j
      - Select sentence with highest similarity if similarity >= t
   c. Add to matches_keep if valid match found

4. For each rollout r in R_remove:
   a. Extract all sentences S_r from rollout r  
   b. Find best matching sentence to target sentence j (same process as step 3b)
   c. Add to matches_remove if valid match found

5. Calculate match rates:
   match_rate_keep = |matches_keep| / |R_keep|
   match_rate_remove = |matches_remove| / |R_remove|
   
6. Return importance(i -> j) = match_rate_keep - match_rate_remove
\end{verbatim}

\subsection{Correlations with the resampling-based importance matrix}

The attention-suppression matrix values correlate with those of the resampling-method matrix. Specifically, the two matrices were positively correlated for 90\% of reasoning traces (mean $r = .20$, 95\% CI: [.12, .27]). This association is stronger when considering only cases fewer than five sentences apart in the reasoning trace, which may better track direct rather than indirect effects represented by the resampling method (mean $r = .34$ [.27, .40]). The magnitudes of these correlations are substantial, given that the two measures capture partially different aspects of causality and the resampling measure itself contains stochastic noise. Hence, these results give weight to the validity of the resampling approach, whose precision we leverage for the forthcoming case study.

\end{document}